\documentclass{article}
\usepackage{arxiv}
\usepackage[utf8]{inputenc} 
\usepackage{tabularx}
\usepackage[T1]{fontenc}    
\usepackage{hyperref}       
\usepackage{url}            
\usepackage{booktabs}       
\usepackage{amsfonts}       
\usepackage{nicefrac}       
\usepackage{microtype}      
\usepackage{graphicx}
\usepackage{placeins}       
\usepackage{enumitem}
\usepackage{amsmath}

\usepackage{adjustbox} 
\usepackage{longtable} 

\title{CHIGLU: A Modular Hardware for Stepper Motorized Quadruped Robot — Design, Analysis, Fabrication, and Validation\\}

\author{ \href{https://orcid.org/0009-0008-8854-1878}{\includegraphics[scale=0.06]{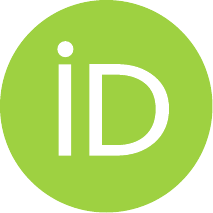}\hspace{1mm}Abid Shahriar}\\
	Department of Electrical and Electronic Engineering, \\
	American International University-Bangladesh\\
	Dhaka-1229, Bangladesh\\
	\href{mailto:abidshahriar97@gmail.com}{abidshahriar97@gmail.com} \\
	 \AND \href{https://orcid.org/0009-0003-3658-6226}{\includegraphics[scale=0.06]{orcid.pdf}\hspace{1mm}Monim Hasan Anik}\\
	 Monim Hasan Anik \\
Department of Electrical and Electronic Engineering,\\
University of Dhaka, Bangladesh\\
\href{mailto:monimhasan-2014416525@eee.du.ac.bd}{monimhasan-2014416525@eee.du.ac.bd} \\
}

\renewcommand{\shorttitle}{\textit{CHIGLU: A Modular Hardware for Stepper Motorized Quadruped Robot — Design, Analysis, Fabrication, and Validation}}

\hypersetup{
    pdftitle={CHIGLU: A Modular Hardware for Stepper Motorized Quadruped Robot — Design, Analysis, Fabrication, and Validation},
    pdfsubject={Robotics Engineering, Hardware Design, Stepper Motors},
    pdfauthor={Abid Shahriar, Monim Hasan Anik},
    pdfkeywords={Modular Hardware, Stepper Motor, Quadruped Robot, PCB Design, PDN Analysis},
}

\begin{document}
\maketitle

\begin{abstract}
	Bio-engineered robots are under rapid development due to their maneuver ability through uneven surfaces. This advancement paves the way for experimenting with versatile electrical system developments with various motors. In this research paper, we present a design, fabrication and analysis of a versatile Printed Circuit Board (PCB) as the main system that allows for the control of twelve stepper motors by stacking low-budget stepper motor controller and widely used Micro-controller unit. The primary motivation behind the design is to offer a compact and efficient hardware solution for controlling multiple stepper motors of a quadruped robot while meeting the required power budget. The research focuses on the hardware’s architecture, stackable design, power budget planning and a thorough analysis. Additionally, PDN (Power Distribution Network) analysis simulation is done to ensure that the voltage and current density are within the expected parameters. Also, the hardware design deep dives into Design for manufacturability (DFM). The ability to stack the controllers on the development board provides insights into the board's components swapping feasibility. The findings from this research make a significant contribution to the advancement of stepper motor control systems of multi-axis applications for bio-inspired robot offering a convenient form factor and a reliable performance. 
\end{abstract}


\begin{table}[htbp]
\centering
\caption{Specifications Table}
\begin{tabular}{ll}
\hline
\textbf{Hardware name} & CHIGLU Modular Hardware for Stepper Motorized Quadruped Robot. \\
\hline
\textbf{Subject area} & Robotics Engineering \\
\hline
\textbf{Hardware type} & Modular Hardware as PCB-Motherboard for Stepper Motorized Quadruped Robot. \\
\hline
\textbf{Closest commercial analog} & No commercial analog is available \\
\hline
\textbf{Open-source license} & CERN OHL \\
\hline
\textbf{Cost of hardware} & 70.93 USD \\
\hline
\textbf{Source file repository} & 
\href{https://doi.org/10.17632/zzxhyjs7pt.2}{https://doi.org/10.17632/zzxhyjs7pt.2} \\
\hline
\end{tabular}
\end{table}

\section{Introduction}

In recent years, quadruped robots have garnered significant attention from researchers and developers due to their ability to traverse rough terrains with greater energy efficiency compared to wheel-based robots. Among the various studies conducted on quadruped robots, numerous researchers have focused on the design and development of different control systems. Depending on the type of electrical drive, diverse controller boards have been created by these researchers \cite {ref1,ref2,ref3} .

HyperDog is a newly developed quadruped robot built on the ROS (Robot Operating System) platform, featuring servo motors as actuators for the legs and an STM32 microcontroller \cite{ref4}. The research briefly discusses the electrical system, highlighting its simplicity attributed to the low power demands of the servo motors. Consequently, the circuit board design is simple as there is no high voltage power needed for the use case of servo motors and does not necessitate complex calculations.

The most suitable approach to building a quadruped robot involves using brushless DC (BLDC) motors, due to its enormous torque holding capacity and advanced control techniques \cite{ref5, ref6, ref7, ref8}. A researcher from New York University presented an open-source torque-controlled legged robot system to teach robotics to high school interns. In this project \cite{ref9}. They have used multiple brushless motor drivers. However, there is not any publicly available information provided on whether they used a full onboard motherboard system to connect all the electrical peripherals or how the electronic design automation was achieved in this scenario \cite{ref9}. Most of the higher-end quadruped built robotic systems utilize brushless motor-based control systems. This is because BLDC (Brushless Direct Current) motors are efficient for high torque and higher power density, which is why these motor types are used to build agile robots like MIT’s Mini Cheetah \cite{ref10}, Zurich’s ANYmal \cite{ref11}, and KAIST HOUND \cite{ref12}. However, there are not enough resources or materials based on quadruped robots with stepper motor built. Even with the widespread use of BLDC motors for their torque and weight-holding capabilities in this use case scenario, research shows that the stepper motors are still a promising option due to their economical cost and the feasibility of using driver ICs, ensuring they remain affordable and accessible \cite{ref13, ref14, ref15, ref16}.

We propose to design and fabricate a motherboard controller for this operation. The controller printed circuit board (PCB) is designed to meet the required power budget (see Fig. \ref{fig:power_budget}). We have also analyzed it with a PDN analyzer for validation and to gain insights in the pre-fabrication stage.

\begin{figure}[h!]
    \centering
    \includegraphics[width=\linewidth]{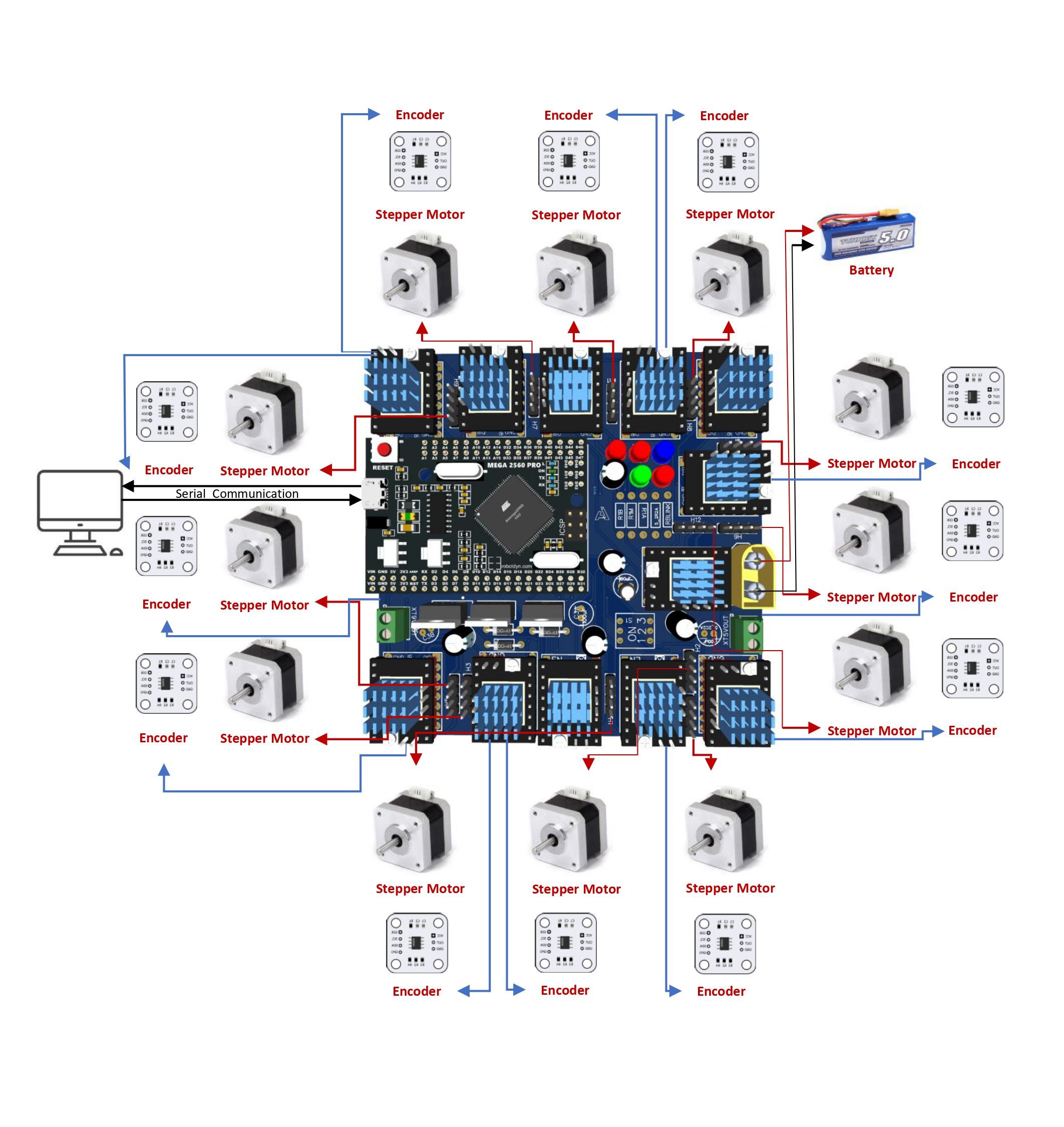}
    \caption{The control scheme of proposed quadruped robot}
    \label{fig:control_scheme}
\end{figure}

\begin{figure}[h]
    \centering
    \includegraphics[width=\linewidth]{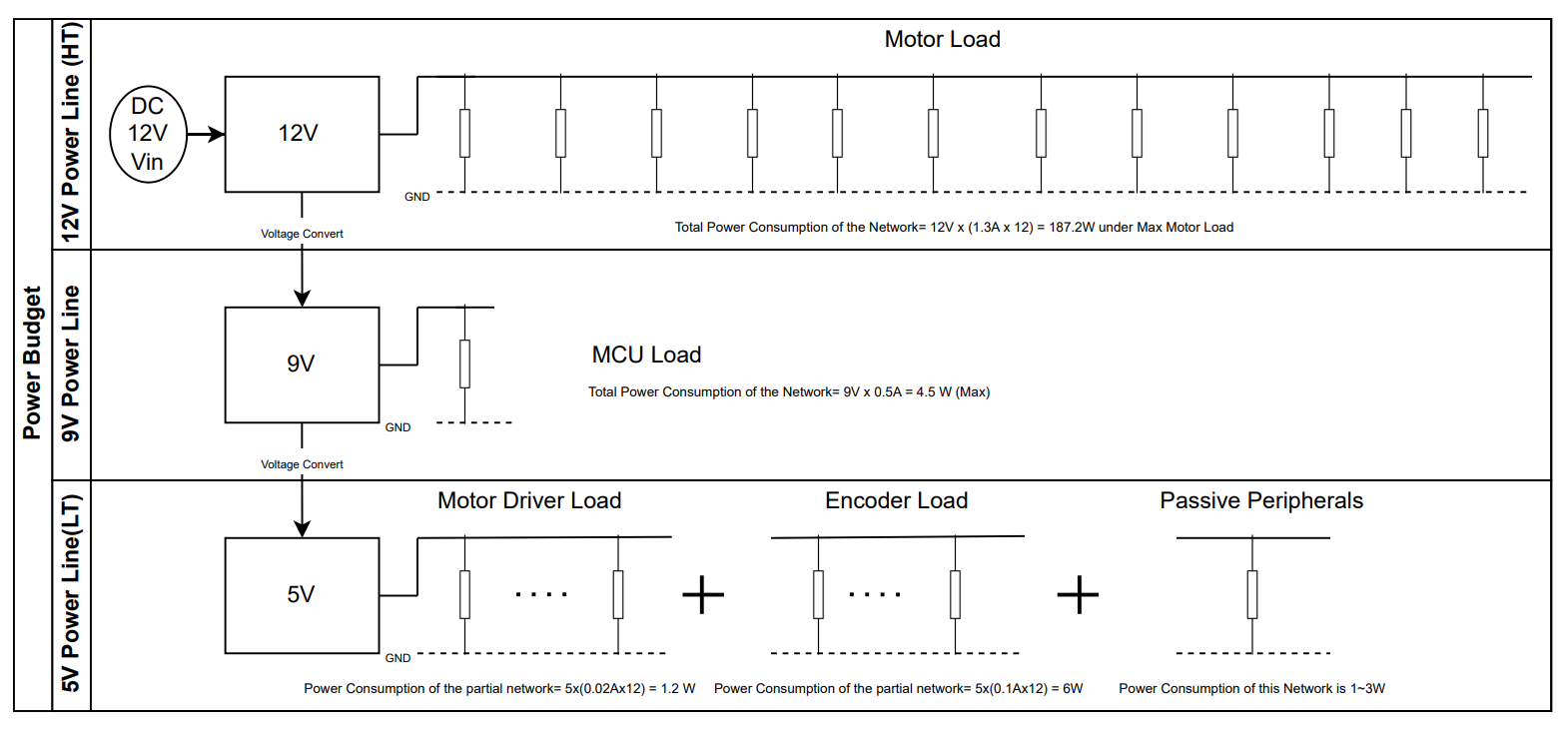}
    \caption{Power Budget Estimation}
    \label{fig:power_budget}
\end{figure}

The controller board is designed focusing on expansion capability, modularity, and compactness, without compromising on the support for essential components (see Fig. \ref{fig:control_scheme}). We accomplished the design requirements in a densely packed two-layer stacked PCB, which has given us a competitive edge in terms of design and cost-effectiveness. The designed board is uniquely small (less than the size of a 6-inch pizza, measured diagonally) to fit into both small and larger-scale quadruped robots' cores. Most of the components on the board are stackable, allowing for open-source development and the ability to quickly hot-swap peripherals. This controller board is suitable for any 12-DoF capable stepper motor-based quadruped robot build.

The control strategy for this system is a distributed design scheme. For controlling each motor, we have chosen the Arduino Mega PRO MCU as the lower control processing unit. The upper control processing unit is a Linux-based computer, which is used for motion planning, posture control, and gait control, and it communicates with the lower computer over serial communication. This work focuses on the lower control unit; hence, signal acquisition and power distribution in this system will be considered. Our motherboard design condenses all required peripherals for control and feedback into an all-in-one solution. The system is designed to take one 12V DC power supply, which can drive 12 DRV/TMC series motor drivers, supply 12V directly to 12 stepper motors, and also power built-in peripherals in what we call a stackable shield board solution. Besides the 12V power signal, we have included two more power signals: 5V to power low-signal peripherals such as feedback encoders, indicator LEDs, and TMC2208 silent motor controllers, and 9V to power the MCU unit. The main challenge of this control board design is how densely we can populate peripheral footprints in both a thermally and electrically efficient manner while keeping the size compact. To ensure proper power and signal transmission through traces, we used a PDN analyzer before designing and fabricating our circuit board. To the best of our knowledge, recent studies and developments of control boards have skipped power density validation, which may cause power deficiencies and improper functioning of the robot.

Our research indicates that developers prefer BLDC motors for building quadruped robots due to their significant torque and speed capabilities \cite{ref15}, \cite{ref16}. However, stepper motor-based quadruped robots can offer more precise positioning compared to BLDC-based designs, as stepper motors utilize micro stepping and can lock into position without drawing additional current \cite{ref13}, a feature not typically found in BLDC systems \cite{ref17}, \cite{ref8}. In our research, we have found very limited options in recent studies in which custom-made motherboards are specifically designed for quadruped robots to power and control multiple stepper motors \cite{ref18} \cite{ref19} \cite{ref20} \cite{ref21} \cite{ref22} \cite{ref23} \cite{ref24} \cite{ref25}. To mitigate this limitation, our study centers on the design, analysis, and fabrication of a motherboard with the necessary functions and capabilities to control a full-fledged 12-DoF stepper motor-based quadruped robot. An additional challenge in this design and fabrication process was to meet the power budget requirements. This open-source board can also serve as a valuable resource for researchers and developers, facilitating its use in various applications.

\FloatBarrier 

\FloatBarrier 

\section{Hardware Description}
We have designed our PCB to meet certain requirements. Salient features of the PCB motherboard include:

\begin{enumerate}
    \item Integration of Arduino Mega PRO mini-Board, which is more compact than Arduino Mega. An SMD-based Arduino Mega MCU shield optimizes space usage.
    \item Compact dimensions, measuring less than the diagonal size of a 6-inch pizza. We concentrated on making the board as compact as possible to ensure this.
    \item Incorporation of inboard chopper functionality for efficient power conversion (12V to 5V \& 9V) utilizing L7805 \& L7809 (see Fig. 2).
    \item Required output current and impedance accuracy for driving 12 stepper motors.
    \item Utilization of twelve AS5600 Encoder Pinouts for angle feedback.
    \item Inclusion of a 3-bit DIP switch and LED indicators for control and status indication.
    \item Implementation of inbuilt reverse current protection for added safety measures.
    \item Meeting the power budget estimation as stated in Fig. 2.
\end{enumerate}

The printed circuit board has been designed using EasyEDA and further analyzed with the PDN Analyzer in Altium Designer. The designed requirement states of controlling a 12-DoF quadruped robot, in which each leg has 3 motors equipped for three degrees of freedom meaning we need at least 24 I/O (Input/Output) pins for the motor signals only. Hence, we need the maximum number of I/O capable MCU (Microcontroller Unit). For this reason, the widely available and cost-efficient ATmega2560-based Arduino Mega MCU is our best choice. The SMD-based Arduino Mega PRO offers the same MCU at a 16MHz clock speed with 54 digital I/O pins, where 15 pins are capable of producing PWM (Pulse Width Modulation) signals. This effectively meets the desired I/O pin requirements within a small form factor.

The power signals carried in this PCB are crucial because the traces must be capable of carrying the large amount of current required for each stepper motor to work properly. To meet this challenging requirement, we have used a PDN analyzer to visually verify and observe each trace’s current density and voltage drop across the board. We then completed the routing of the PCB layout. To meet the demand for the hardware, we have chosen FR4(Flame retardant material) which has a good strength-to-weight ratio and also HASL (Hot Air Solder Leveling) surface finish for the manufacturing. FR4 is known for its flame-retardant properties and robust strength-to-weight ratio which ensures that the PCB can withstand the high currents required by stepper motors, maintaining reliability under load. HASL (Hot air solder leveling) surface finish also enhances solder-ability and ensures robust electrical connections, crucial for efficient power transmission and reliable operation of the PCB hardware.

\begin{table}[htbp]
\centering
\caption{Major Power Net Port and Connected Peripherals}
\begin{tabularx}{\textwidth}{lX X c l}
\toprule
\textbf{NetPort Name} & \textbf{Connected \newline Peripherals} & \textbf{Connected \newline Components \newline Designation} & \textbf{Number of Connected Components} & \textbf{Signal Value} \\
\midrule
LT & TMC2208 Motor Driver (VIO pin), AS5600 Encoder & DRV1, DRV2, DRV3, DRV4, DRV5, DRV6, P1, P2, P3, P4, P5, P6 & 12 & 5V \\
LTX & TMC2208 Motor Driver & DRV7, DRV8, DRV9, DRV10, DRV11, DRV12, P7, P8, P9, P10, P11, P12 & 12 & 5V \\
L9T & Arduino Mega PRO MINI & U2 & 1 & 9V \\
HT & TMC2208 Motor Driver (VM, VMOT PIN), Input Switch, Power Indicator LED & DRV1, DRV2, DRV3, DRV4, DRV5, DRV6, DRV7, DRV8, DRV9, DRV10, DRV11, DRV12, S1, R1M & 14 & 12V \\
\bottomrule
\end{tabularx}
\end{table}

\FloatBarrier 

\begin{figure}[h!]
    \centering
    \includegraphics[width=6.5cm]{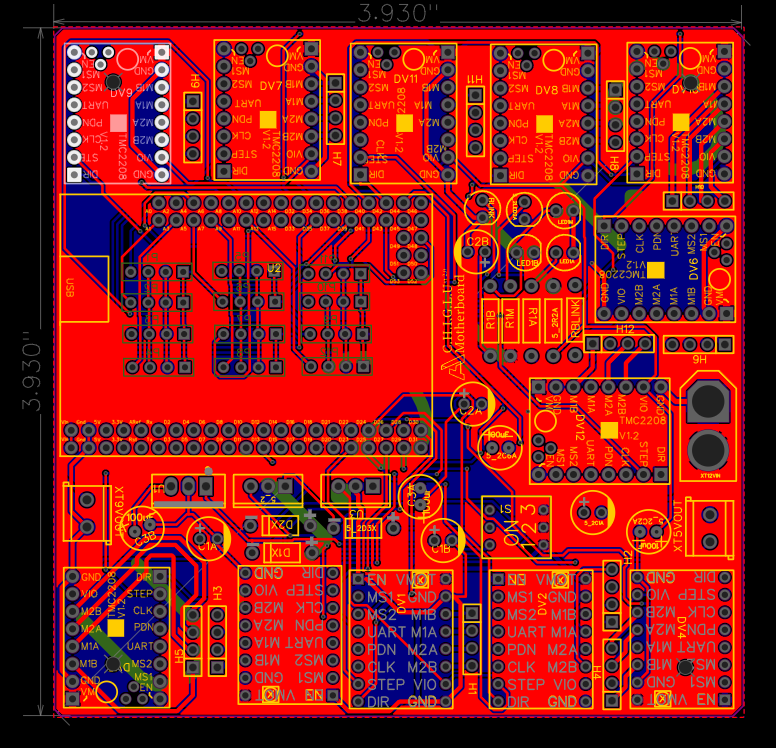}
    \caption{Layout of the Design}
    \label{fig:Layout of the Design}
\end{figure}
\FloatBarrier 

In the layout design, we meticulously routed the tracks, as illustrated in Fig. 3 and Fig. 5. The major NETs are defined in Table 1. Initially, we routed the HT Net, as specified in the schematic. By HT Net, we mean the high-tension voltage trace, which includes the stepper motor's power line and also links to the power converter IC's input. Since this trace draws a maximum amount of current, we have to set the width of this trace to more than 15 mil which we have theoretically calculated in the validation and characterization section to transfer the required amount of current rated by the motor. Later, we will validate this with the PDN analyzer.
We have densely integrated a large number of Net layers during routing. For any traces that are not part of a power plane and are under 12V, we used a width of 10 mil, which is sufficient for running low-powered peripherals. Consequently, we have connected nets labeled LT, LTX, motor drivers, and encoders (see Table 1) and and L9T to the MCU (Micro-Controller Unit).
As depicted in the 3D view (see Fig. 4.), the top layer has all the basic peripherals connected to the top side of the PCB. On the other side of the PCB, we connected the encoder outputs. By doing this, we optimized space and achieved feasibility, while also minimizing PCB printing costs by implementing the maximum number of components within two layers.
\begin{figure}[h!]
    \centering
    \begin{minipage}{0.49\textwidth}
    \centering
    \includegraphics[width=6cm]{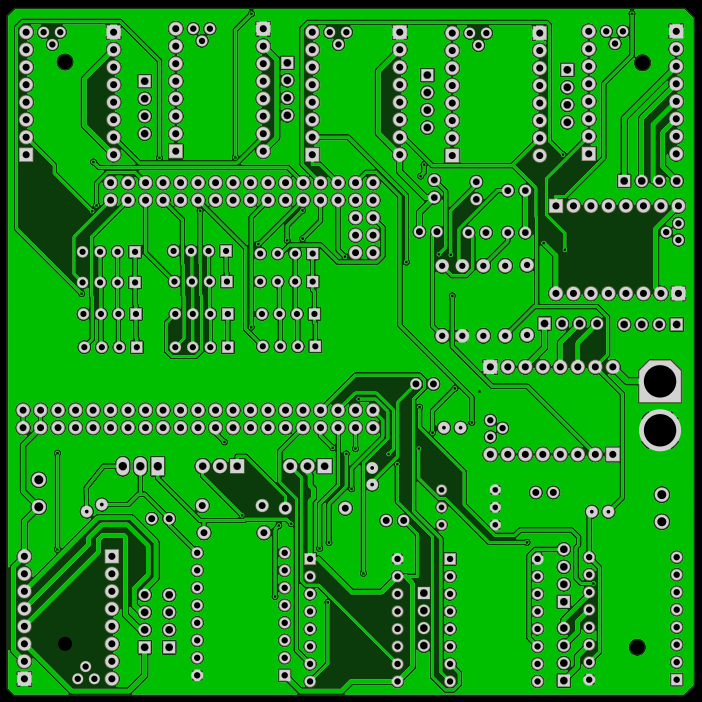}
    \label{fig:}
    \end{minipage}
    \hfill
    \begin{minipage}{0.49\textwidth}
    \centering
    \includegraphics[width=6cm]{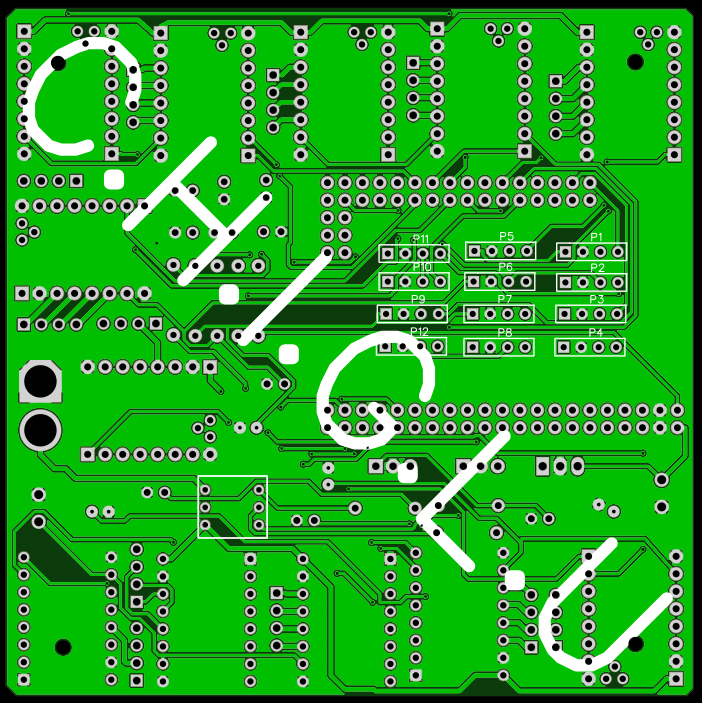}
    \label{fig:3D View (Front and Back) of the Designed Printed Circuit Board}
    \end{minipage}
    \caption{3D View (Front and Back) of the Designed Printed Circuit Board}
\end{figure}
\FloatBarrier 
The circuit board is designed to draw power from a 12V supply, making it compatible with any 12V, 3S LiPo battery that can be connected directly from the battery to the hardware board. To power the Arduino Mega Pro, we chose to convert the 12V power supply to 9V. This is because the Arduino Mega development board's inbuilt power converter will handle the conversion from 9V. We opted to power the Arduino Mega with 9V to ensure stability; in case of any power fluctuations or voltage drops, the Arduino can continue running since we are supplying more than the minimum required voltage of 5V.
\begin{figure}[h!]
    \centering
    \includegraphics[width=7cm]{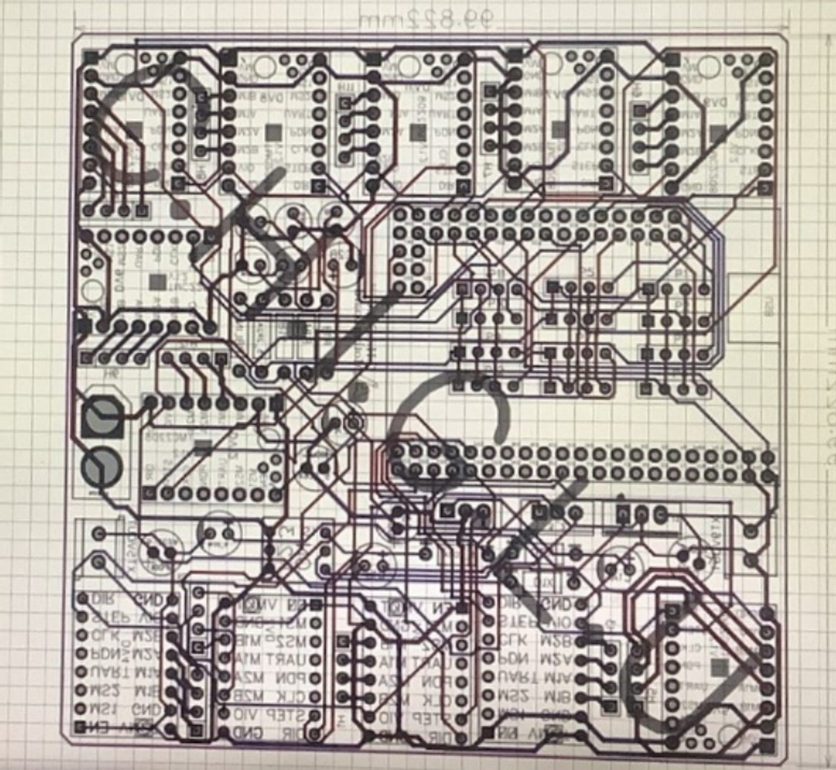}
    \caption{X-ray View}
    \label{fig:X-ray View}
\end{figure}
\FloatBarrier 
In the schematic shown in Fig.13., we have implemented three buck converters using widely available L7805 and L7809 series power MOSFETs to support low-powered peripherals. In Fig. 7 and Fig. 13, these blocks also include LED status indicators for MCU activity, power converter signals, and main power board 
signals. In the schematic (see Fig. 10), we have interfaced the Arduino MCU with connected peripherals, including motor driver signals and analog and digital signals. Additionally, a header pin has been added in the schematic for faster and hot-swappable motor driver stacking (see Fig. 10) and connecting encoders to the board (see Fig. 12).
In the schematic (see Fig. 9) we have connected 12 motor controllers powered in a parallel connection similar to a bus bar. We supplied 5V power from the converted 5V source, as indicated by the LT and LTX Net ports (see Table 1) connected to VIO pins, along with a common ground (GND). To ensure proper power delivery, we utilized two buck 5V circuits for efficient power conversion and transmission to the motor drivers to keep it empowering in case of voltage fluctuation. Hence, we have reinforced two power converters (Netport LT and LTX) to support all the peripherals. We have added 5V and 9V power outputs in the board to extend the board’s power-extending capability (see Figure 8), and also to cut off all the operations and to ensure safety we have also implemented kill switch (power switch) (see Figure 6) and a LED to see the observe the status when uploading code (see Figure 7). The enable pin in the schematic (see Figure 9) can be either set to active high or active low, for our purpose we have set all the drivers enable pin to active low.

Now, the biggest challenge is how to densely populate these peripherals and accommodate these footprints to meet the power budget (on a small-scale PCB. We didn't use any SMT components, even though they would have saved space. Instead, for ease of soldering and the open source and easy DIY (do it yourself) feasibility we have opted for a full through-hole component design. The routing intricacies of this design involve sophisticated DRC (Design rule check) to achieve such complex routing design architecture. We have kept a minimum track width of 15mil, clearance of 10mil, via diameter of 24mil, and via drill 12mil minimum width in routing while considering clearance, pad-to-pad distance, thermal integrity, pad-to-track distance and ensuring proper impedance through the track lines. 
\FloatBarrier 
\begin{figure}[h!]
    \centering
    \includegraphics[width=7cm]{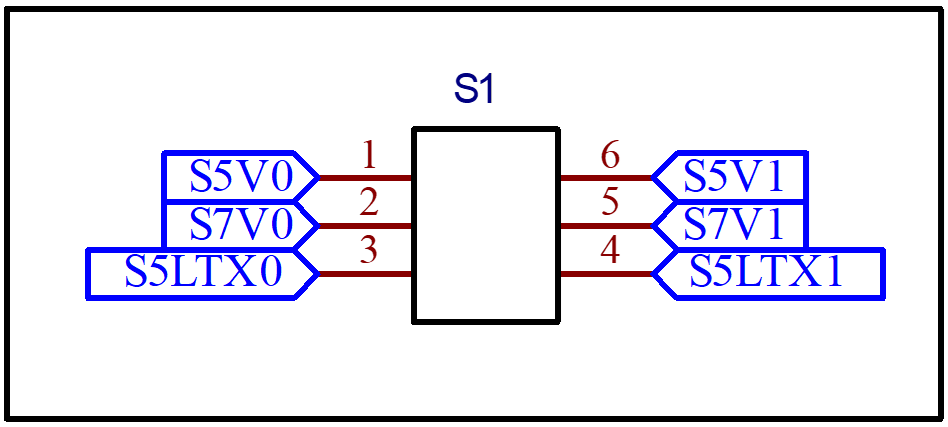}
    \caption{Power Switch}
    \label{fig:Power Switch}
\end{figure}
\FloatBarrier 

\begin{figure}[h!]
    \centering
    \includegraphics[width=7cm]{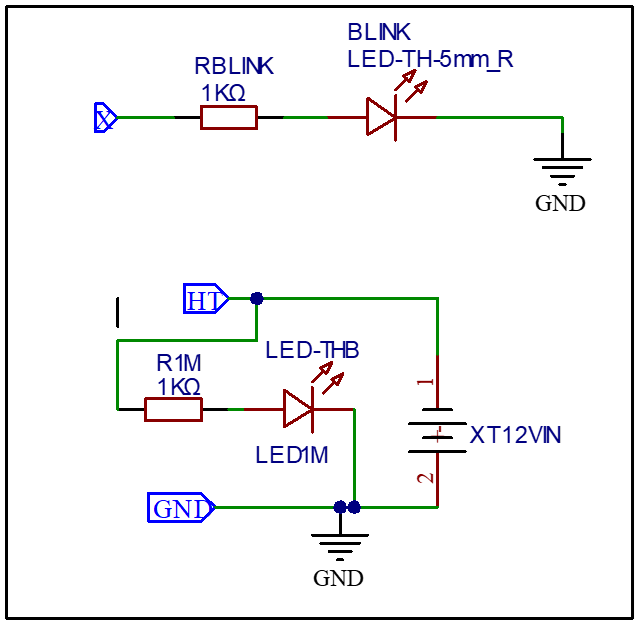}
    \caption{Power Input connector (DC 12VIN) and Indicator LED}
    \label{fig:Power Input connector (DC 12VIN) and Indicator LED}
\end{figure}
\FloatBarrier 

\begin{figure}[h!]
    \centering
    \includegraphics[width=10cm]{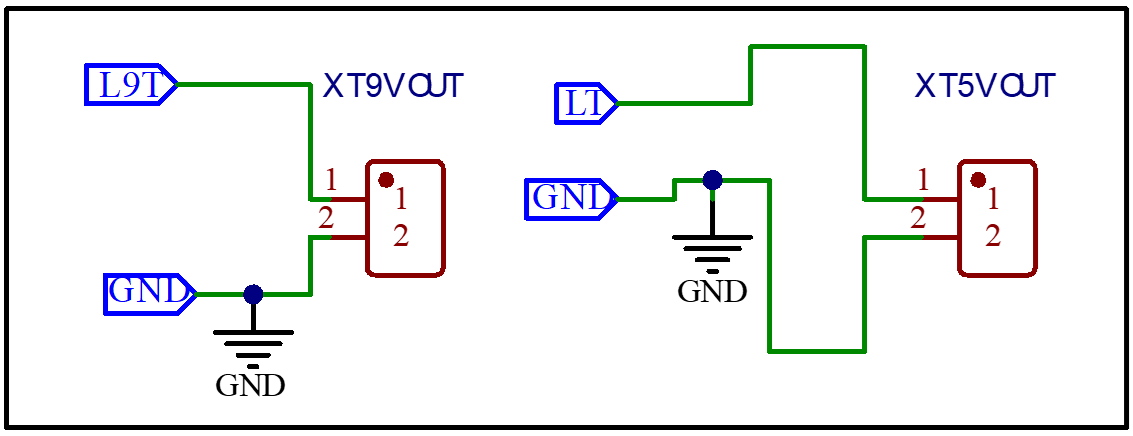}
    \caption{Board Power Extension Output (9V and 5V) Schematic}
    \label{fig:Board Power Extension Output (9V and 5V) Schematic}
\end{figure}
\FloatBarrier 

\begin{figure}[h!]
    \centering
    \includegraphics[width=15cm]{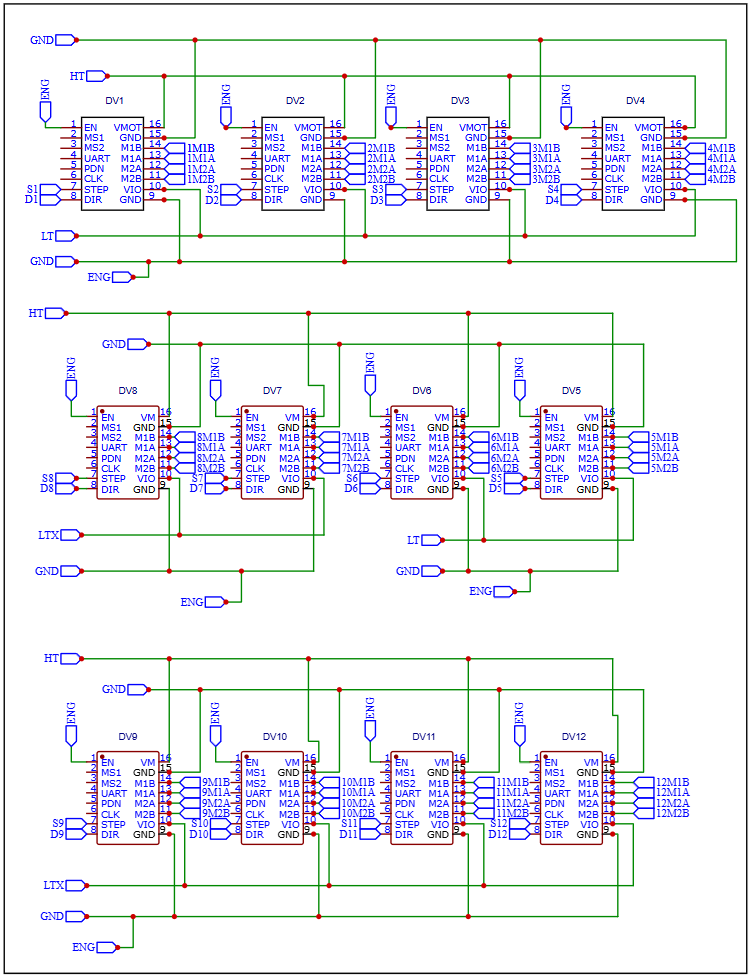}
    \caption{Motor Driver Schematic}
    \label{fig:Motor Driver Schematic}
\end{figure}
\FloatBarrier 

\begin{figure}[h!]
    \centering
    \includegraphics[width=10cm]{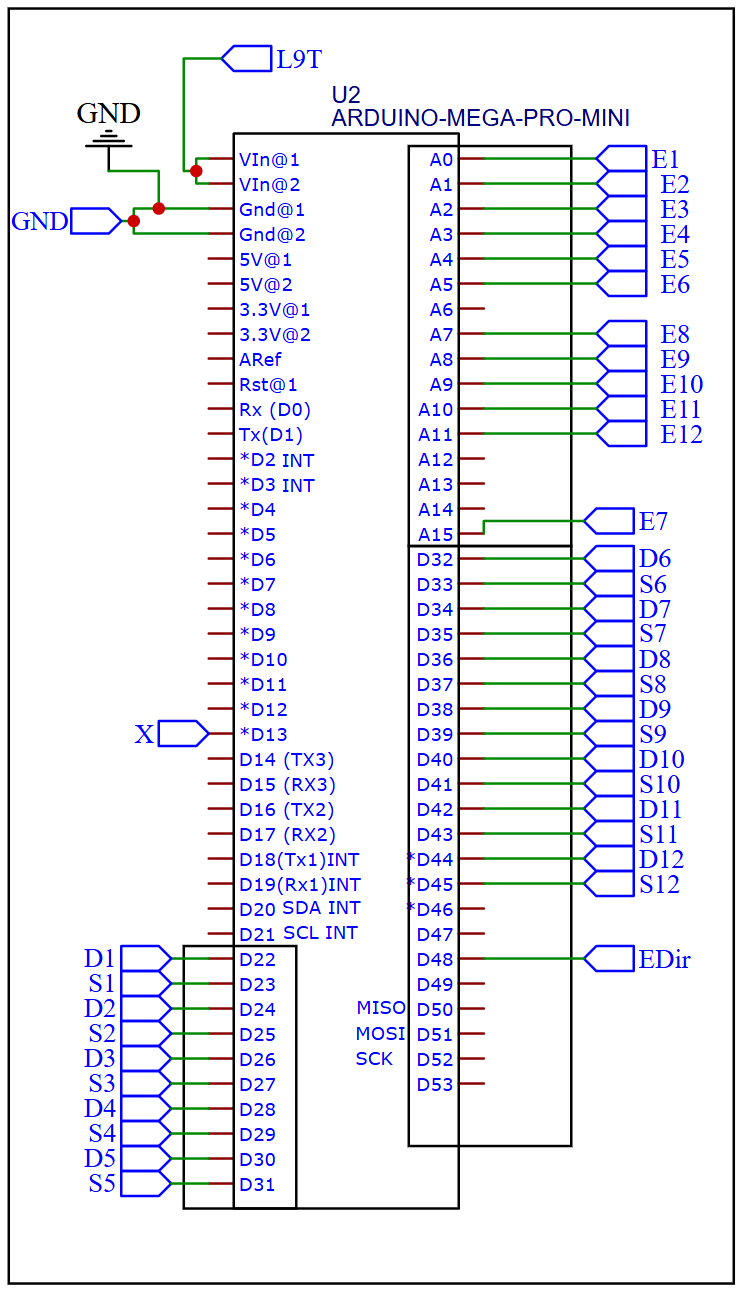}
    \caption{Microcontroller Unit Schematic}
    \label{fig:Microcontroller Unit Schematic}
\end{figure}
\FloatBarrier 

\begin{figure}[h!]
    \centering
    \includegraphics[width=11.5cm]{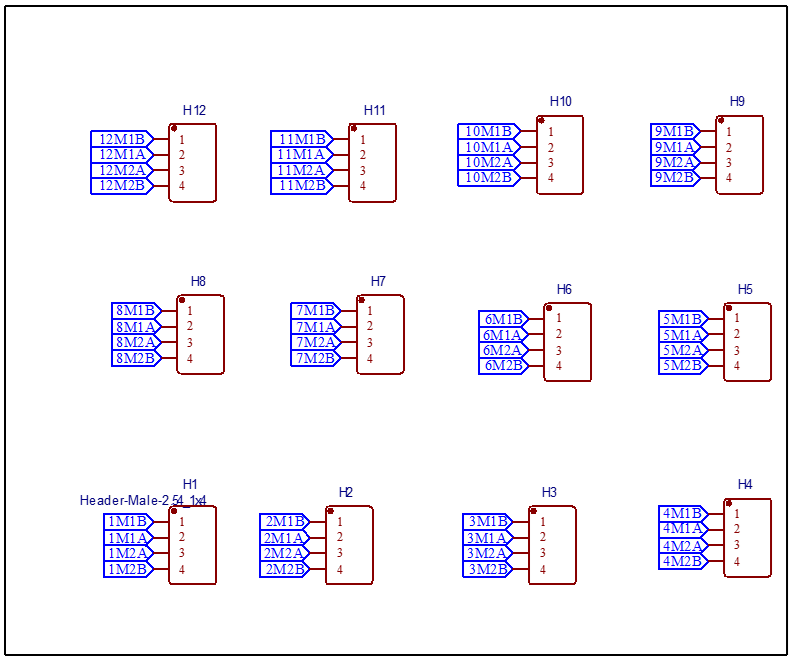}
    \caption{Stepper Motor Output Schematic}
    \label{fig:Stepper Motor Output Schematic}
\end{figure}
\FloatBarrier 

\begin{figure}[h!]
    \centering
    \includegraphics[width=11.5cm]{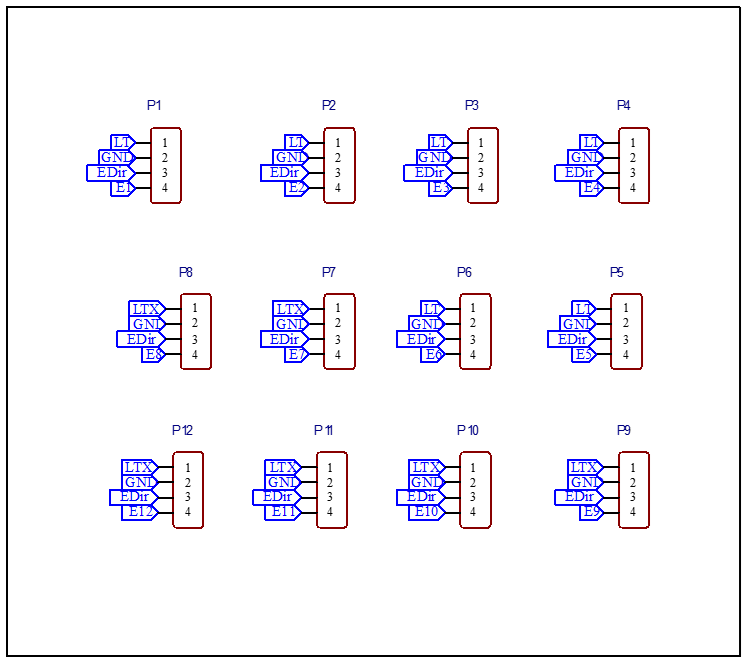}
    \caption{Encoder Signal Schematic}
    \label{fig:Encoder Signal Schematic}
\end{figure}
\FloatBarrier 
\hfill \break
\hfill \break
\hfill \break

\begin{figure}[h!]
    \centering
    \includegraphics[width=15cm]{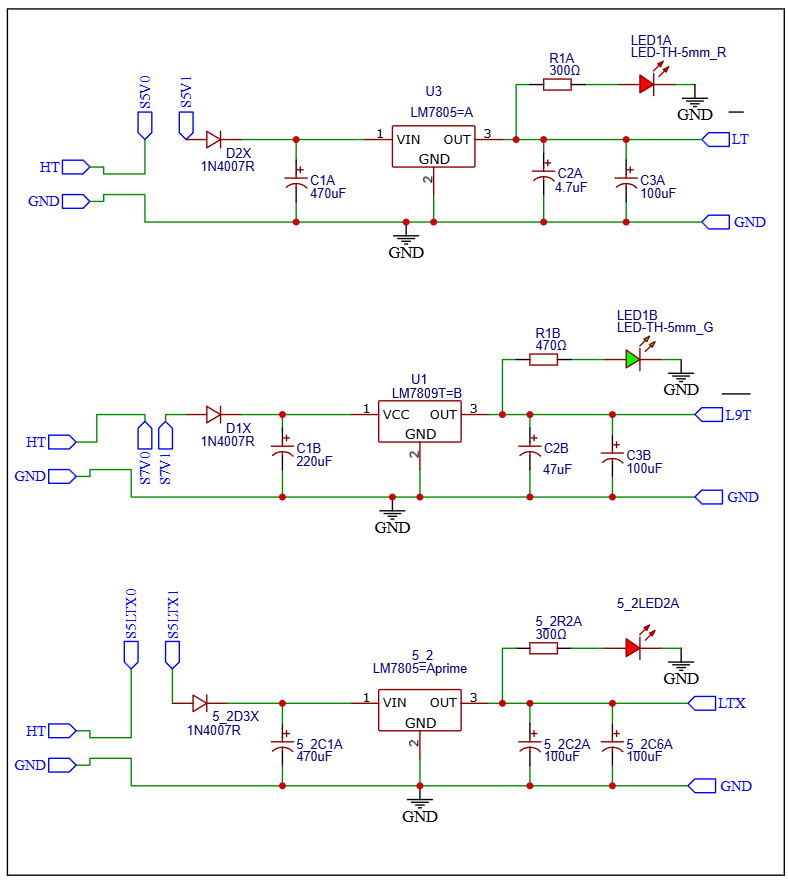}
    \caption{Power Converters Schematic}
    \label{fig:Power Converters Schematic}
\end{figure}
\FloatBarrier 
\hfill \break
\hfill \break
\hfill \break
\hfill \break
\hfill \break
\hfill \break
\section{Design files summary}

\begin{table}[h!]
\centering
\caption{Design Files Information}
\begin{tabularx}{\textwidth}{ X X X X }
\hline
\textbf{Design file name} & \textbf{File type} & \textbf{Open-source license} & \textbf{Location of the file} \\ \hline
BOM.xlsx & Excel & CERN OHL & \url{https://doi.org/10.17632/zzxhyjs7pt.2} \\ \hline
Box\_Assembly.20240604-205600.FCBak & FCBAK File & CERN OHL & \url{https://doi.org/10.17632/zzxhyjs7pt.2} \\ \hline
Box\_Assembly.FCStd & FreeCAD Document & CERN OHL & \url{https://doi.org/10.17632/zzxhyjs7pt.2} \\ \hline
Enclosure.20240604-210120.FCBak & FCBAK File & CERN OHL & \url{https://doi.org/10.17632/zzxhyjs7pt.2} \\ \hline
Enclosure.FCStd & FreeCAD Document & CERN OHL & \url{https://doi.org/10.17632/zzxhyjs7pt.2} \\ \hline
Enclosure\_lid.20240604-210120.FCBak & FCBAK File & CERN OHL & \url{https://doi.org/10.17632/zzxhyjs7pt.2} \\ \hline
Enclosure\_lid.FCStd & FreeCAD Document & CERN OHL & \url{https://doi.org/10.17632/zzxhyjs7pt.2} \\ \hline
PCB\_enclosure\_final. \newline 20240604-212250.FCBak & FCBAK File & CERN OHL & \url{https://doi.org/10.17632/zzxhyjs7pt.2} \\ \hline
PCB\_enclosure\_final.FCStd & FreeCAD Document & CERN OHL & \url{https://doi.org/10.17632/zzxhyjs7pt.2} \\ \hline
Automatic Optical Imaging Data.rar & WinRAR Archive & CERN OHL & \url{https://doi.org/10.17632/zzxhyjs7pt.2} \\ \hline
CAM.PNG & PNG File & CERN OHL & \url{https://doi.org/10.17632/zzxhyjs7pt.2} \\ \hline
CAM\_bot.dpf & DPF File & CERN OHL & \url{https://doi.org/10.17632/zzxhyjs7pt.2} \\ \hline
CAM\_drill.dpf & DPF File & CERN OHL & \url{https://doi.org/10.17632/zzxhyjs7pt.2} \\ \hline
CAM\_maskb.dpf & DPF File & CERN OHL & \url{https://doi.org/10.17632/zzxhyjs7pt.2} \\ \hline
CAM\_maskt.dpf & DPF File & CERN OHL & \url{https://doi.org/10.17632/zzxhyjs7pt.2} \\ \hline
CAM\_outline.dpf & DPF File & CERN OHL & \url{https://doi.org/10.17632/zzxhyjs7pt.2} \\ \hline
CAM\_silkb.dpf & DPF File & CERN OHL & \url{https://doi.org/10.17632/zzxhyjs7pt.2} \\ \hline
CAM\_silkt.dpf & DPF File & CERN OHL & \url{https://doi.org/10.17632/zzxhyjs7pt.2} \\ \hline
CAM\_top.dpf & DPF File & CERN OHL & \url{https://doi.org/10.17632/zzxhyjs7pt.2} \\ \hline
Drill Gcode.ex1 & Executable File & CERN OHL & \url{https://doi.org/10.17632/zzxhyjs7pt.2} \\ \hline
Layout Altium \newline version.pcbdoc & Altium PCB Document & CERN OHL & \url{https://doi.org/10.17632/zzxhyjs7pt.2} \\ \hline
rout Gcode.ex1 & Executable file & CERN OHL & \url{https://doi.org/10.17632/zzxhyjs7pt.2} \\ \hline
Schematic EasyEDA \newline version.json & JavaScript Object Notation File & CERN OHL & \url{https://doi.org/10.17632/zzxhyjs7pt.2} \\ \hline
Schematic Altium \newline version.schdoc & Schematic document & CERN OHL & \url{https://doi.org/10.17632/zzxhyjs7pt.2} \\ \hline
Layout EasyEDA\newline version.json & JavaScript Object Notation File & CERN OHL & \url{https://doi.org/10.17632/zzxhyjs7pt.2} \\ \hline
PDN Analyzer Report & Directory containing necessary PDN Analysis files & CERN OHL & \url{https://doi.org/10.17632/zzxhyjs7pt.2} \\ \hline
\end{tabularx}
\end{table}

\FloatBarrier 

\begin{itemize}
  \item \texttt{BOM.xlsx} - A spreadsheet file containing a bill of materials for a project or product, detailing components, quantities, and specifications.
  \item \texttt{Box\_Assembly.20240604-205600.FCBak} - A backup file of enclosure assembly project containing saved configurations.
  \item \texttt{Box\_Assembly.FCStd} - A CAD file in FreeCAD format, representing assembly of the enclosure structure.
  \item \texttt{Enclosure.20240604-210120.FCBak} - A backup file of the enclosure containing saved configurations.
  \item \texttt{Enclosure.FCStd} - A FreeCAD file representing assembly of the enclosure.
  \item \texttt{Enclosure\_lid.20240604-210120.FCBak} - A backup file of the enclosure lid containing saved configurations.
  \item \texttt{Enclosure\_lid.FCStd} - A FreeCAD file representing assembly of the enclosure lid.
  \item \texttt{PCB\_enclosure\_final.20240604-212250.FCBak} - A backup file merging the enclosure with the lid for the enclosure assembly project, storing saved configurations.
  \item \texttt{PCB\_enclosure\_final.FCStd} - A FreeCAD file representing assembly of the enclosure (enclosure lid included).
  \item \texttt{Automatic Optical Imaging Data.rar} - AOI (Automatic Optical Imaging Layer).
  \item \texttt{CAM.PNG} - A png file showing layers in CAM process.
  \item \texttt{CAM\_bot.dpf} - Manufacturing file: PCB bottom layer.
  \item \texttt{CAM\_drill.dpf} - Manufacturing file: Drill layer.
  \item \texttt{CAM\_maskb.dpf} - Manufacturing file: PCB bottom-mask layer.
  \item \texttt{CAM\_maskt.dpf} - Manufacturing file: PCB top-mask layer.
  \item \texttt{CAM\_outline.dpf} - Manufacturing file: PCB's outline.
  \item \texttt{CAM\_silkb.dpf} - Manufacturing file: PCB bottom legend print layer.
  \item \texttt{CAM\_silkt.dpf} - Manufacturing file: PCB top Legend Print layer.
  \item \texttt{CAM\_top.dpf} - Manufacturing file: PCB top layer.
  \item \texttt{Drill Gcode.ex1} - A G-code file used for CNC drilling operations. It contains instructions for the machine to drill holes at specified locations and depths in a material.
  \item \texttt{Layout Altium version.pcbdoc} - A PCB layout document created using Altium Designer. It contains detailed information about the placement of components and routing of electrical connections on a printed circuit board.
  \item \texttt{rout Gcode.ex1} - A G-code file used for CNC machining or 3D printing. It contains instructions for the machine to execute specific movements and operations to create a designed part or pattern.
  \item \texttt{Schematic EasyEDA version.json} - A JSON-format document created using EasyEDA, an online EDA tool. It contains schematic information for an electronic circuit, detailing component connections and attributes in a structured JSON format.
  \item \texttt{Schematic Altium version.schdoc} - A schematic document created with Altium Designer. It contains the circuit diagram and details the electronic components and their interconnections for a specific design.
  \item \texttt{Layout EasyEDA version.json} - A JSON-format document created using EasyEDA. It contains layout information for a PCB design, detailing component placements and routing paths in a structured JSON format.
\end{itemize}

\FloatBarrier 

\section{Bill of materials summary}
The Bill of materials file includes the cost of the bills of all the components used in this design and is available here: \url{https://data.mendeley.com/datasets/zzxhyjs7pt/2}

\section{Build Instructions}

\subsection*{PCB Assembly}

\textbf{Step 1: Fabricate the Circuit Board}
\begin{itemize}[label=-]
    \item  Use the provided Gerber files to fabricate the circuit board.
    \item For feasibility, we have provided the CAM (Computer Aided Manufacturing) file which can be used to fabricate the PCB.
\end{itemize}

\textbf{Step 2: Populate the Circuit Board with Components}
\begin{itemize}[label=-]
    \item Populate the circuit board with components as described with designators in the 'Bill of Summary'.
    \item Refer to the silk layer of the Gerber files and the fabricated PCB for information on component placement.
    \item Since all components are through holes, there is no need for a reflow oven; any soldering iron can be used.
\end{itemize}

\textbf{Step 3: Soldering Components}
\begin{itemize}[label=-]
    \item Populate the board with all peripherals before placing the female headers.
    \item After placing other components, proceed to solder the female headers.
\end{itemize}

\textbf{Step 4: Voltage Testing}
\begin{itemize}[label=-]
    \item After soldering, use a voltmeter to check the voltage at different test points:
    \begin{itemize}[label=--]
        \item Motor Controller Input Voltage:
        \begin{itemize}[label=---]
            \item Test Points: Vin, GND
            \item Expected Voltage: 12V
        \end{itemize}
        \item Motor Controller Output Voltage:
        \begin{itemize}[label=---]
            \item Test Points: VM, GND
            \item Expected Voltage: 12V
        \end{itemize}
        \item Converter Voltage Output:
        \begin{itemize}[label=---]
            \item Test Points: LT, L9T, GND
            \item Expected Voltage: 5V
        \end{itemize}
        \item Encoder Voltage Output:
        \begin{itemize}[label=---]
            \item Test Points: LT, GND
            \item Expected Voltage: 5V
        \end{itemize}
    \end{itemize}
    \item Ensure the PCB is correctly fabricated if the test points provide the expected voltages.
\end{itemize}

\textbf{Step 5: Continuity Testing}
\begin{itemize}[label=-]
    \item Test closely assigned annular ring pads for continuity to ensure there are no short connections between pads.
    \item Pay special attention to the MCU pins, which are through holes but closely spaced.
    \item Verify that these pads, connected to various peripherals as signal pins, are not accidentally bridged during soldering.
    \item This step can save time during debugging in case of errors.
\end{itemize}

\textbf{Step 6: Header Alignment}
\begin{itemize}[label=-]
    \item Ensure that the header pins are closely aligned and not bent while soldering.
    \item This ensures the pins sit straight when stacking components.
\end{itemize}

\textbf{Step 7: Stack Peripherals}
\begin{itemize}[label=-]
    \item Stack all peripherals onto the board, including:
    \begin{itemize}[label=--]
        \item Arduino Mega PRO
        \item TMC2208 motor Drivers
        \item AS5600 Encoders
        \item All motor connectors
    \end{itemize}
\end{itemize}

\subsection*{PCB Enclosure Design}

\textbf{Step-by-Step Process for Designing a PCB Enclosure:}

\textbf{Step 1: Define Requirements}
\begin{itemize}[label=-]
    \item Determine the dimensions of the PCB.
    \item Identify the locations of connectors, buttons, and other external interfaces.
\end{itemize}

\textbf{Step 2: Initial Sketch and Conceptualization}
\begin{itemize}[label=-]
    \item Draw a rough sketch of the enclosure design.
    \item Include openings for connectors, buttons, and vents for cooling if necessary.
\end{itemize}

\textbf{Step 3: Choose Software Tools}
\begin{itemize}[label=-]
    \item Select appropriate CAD software for designing the enclosure (FreeCAD was used to design the enclosure).
\end{itemize}

\textbf{Step 4: Create Base Design}
\begin{itemize}[label=-]
    \item Start with a simple box design that fits the dimensions of the PCB.
    \item Add mounting points for securing the PCB inside the enclosure.
\end{itemize}

\textbf{Step 5: Refine the Design}
\begin{itemize}[label=-]
    \item Add detailed features such as holes for screws, cutouts for connectors, and support structures for components.
    \item Ensure there is enough space for wiring and other components that will be inside the enclosure.
\end{itemize}

\textbf{Step 6: Consider Heat Dissipation}
\begin{itemize}[label=-]
    \item Design ventilation openings.
\end{itemize}

\textbf{Step 7: Design Assembly Features}
\begin{itemize}[label=-]
    \item Add features that allow the enclosure to be easily assembled and disassembled (e.g., screws).
\end{itemize}

\textbf{Step 8: Prototype and Test Fit}
\begin{itemize}[label=-]
    \item Create a 3D-printed prototype of the enclosure.
    \item Test fit the PCB and make sure all components align correctly with the enclosure features.
\end{itemize}

\textbf{Step 9: Iterate and Improve}
\begin{itemize}[label=-]
    \item Make necessary adjustments based on the test fit.
    \item Iterate the design until it meets all requirements.
\end{itemize}

\textbf{Step 10: Finalize Design}
\begin{itemize}[label=-]
    \item Finalize the CAD model with all the refined details.
    \item Double-check all dimensions and fitment.
\end{itemize}

\textbf{Step 11: Production Preparation}
\begin{itemize}[label=-]
    \item Prepare the design files for production.
    \item Choose a manufacturing method (e.g., 3D printing, injection molding).
\end{itemize}

\begin{figure}[h!]
    \centering
    \includegraphics[width=7cm]{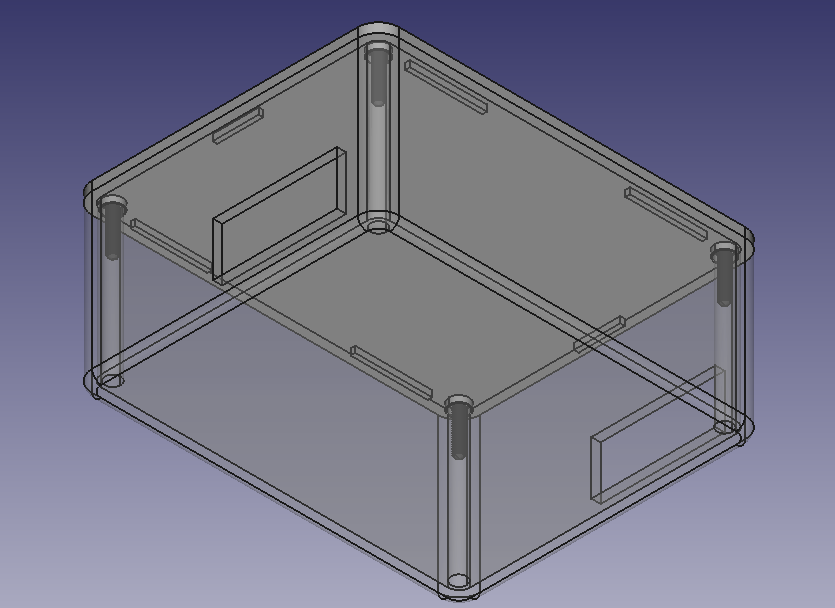}
    \caption{3D Model of PCB Enclosure}
    \label{fig:3D Model of PCB Enclosure}
\end{figure}
\FloatBarrier 

\section{Operation Instructions}
The operating instructions for this hardware are straightforward. The MCU is connected to a computer via USB for serial communication, allowing signals to be sent to the MCU for control purposes. In our initial attempt, we created a only copper processed prototype board which is exactly the same essence as the original design (see Fig. 15.) following the schematic for demonstration purposes as prototype PCB hardware. On this board, we stacked the MCU and one motor driver board with a stepper motor to verify the basic functionality of the board. The entire board is powered by an external 12V power supply, which also drives the motors. The Final PCB board has been assembled (see Fig. 16.) and bare-board test has been done on it and it ensures us the functionality and we powered it up with 12V supply and the board shows indication of life as seen the indicator light glows up. Bare board testing (BBT) and functionality check is done extensively on the final board in section 7: Hardware testing and validation of this study. 

Initially, we launch the Arduino IDE on our computer and upload a sketch designed to receive angles. The angles are calculated by an inverse kinematics script developed from prior research \cite{ref27}. This script computes the angles and transmits them to the Arduino via serial communication. Stepper motors are known for their exceptional precision compared to other types of motors. To enhance this precision, we utilize an AS5600 encoder to encode and transmit angle data to the Arduino. We also upload an Arduino sketch to the microcontroller, assigning the necessary pins to the corresponding TMC2208 motor driver. The computed angles are then sent to the stepper motor, while the actual step angle information from the encoder is received and monitored via the serial monitor. The detailed flowchart of the working principle is shown in the flowchart (see Fig. 17.). The Micro controller runs and keeps executing the process as long as the serial communication is intact. 
\begin{figure}[h!]
    \centering
    \begin{minipage}{0.49\textwidth}
    \centering
    \includegraphics[width=6cm]{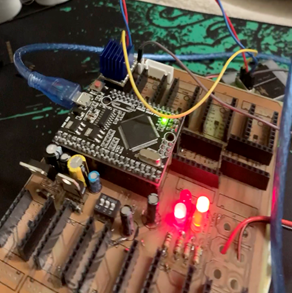}
    \label{fig:Bareboard 1}
    \end{minipage}
    \hfill
    \begin{minipage}{0.49\textwidth}
    \centering
    \includegraphics[width=6cm]{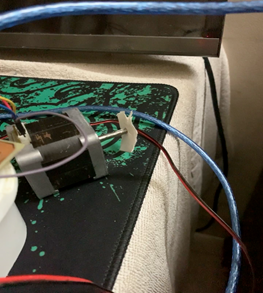}
    \label{fig:Bareboard 2}
    \end{minipage}
    \caption{Prototype board stacked with MCU, one motor driver board with a stepper motor and external 12V power supply}
\end{figure}
\FloatBarrier 

\begin{figure}[h!]
    \centering
    \includegraphics[width=10cm]{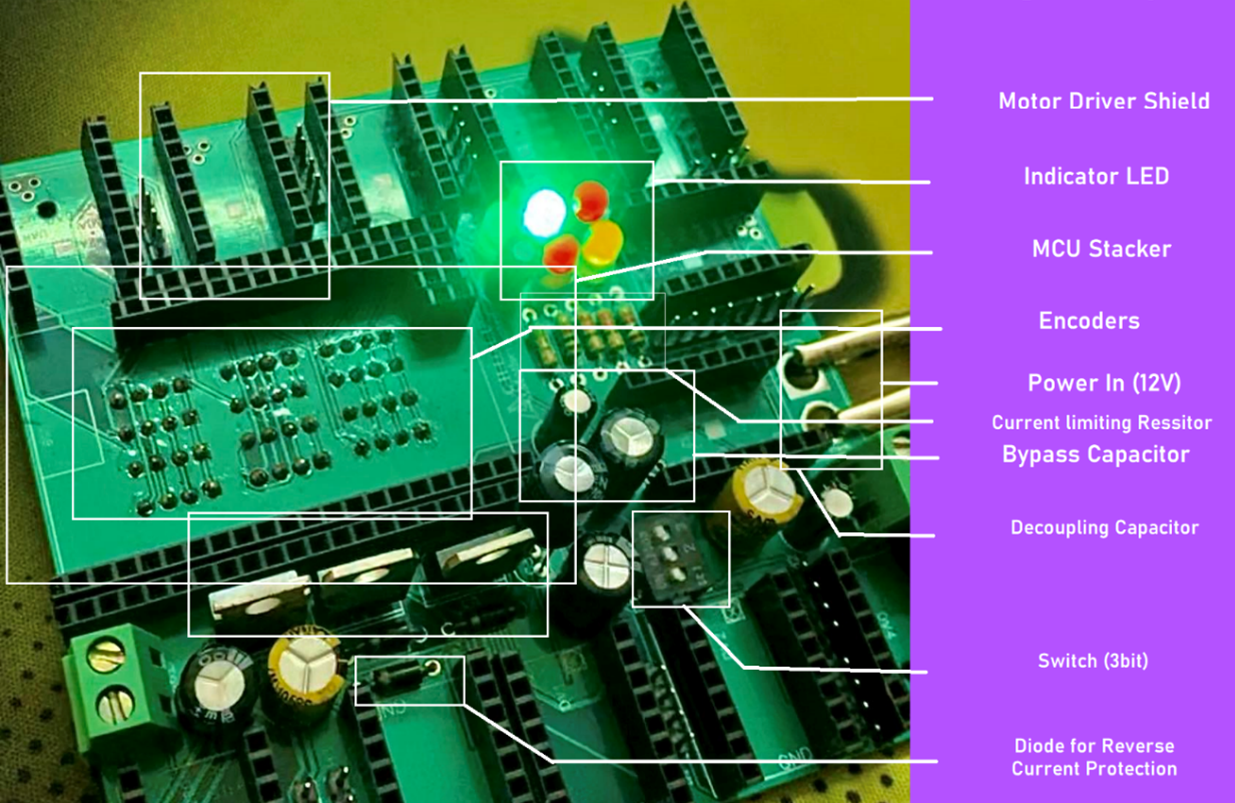}
    \caption{Completed Assembly of the Final PCB Board}
    \label{fig:Completed Assembly of the Final PCB Board}
\end{figure}
\FloatBarrier 

\begin{figure}[h!]
    \centering
    \includegraphics[width=13cm]{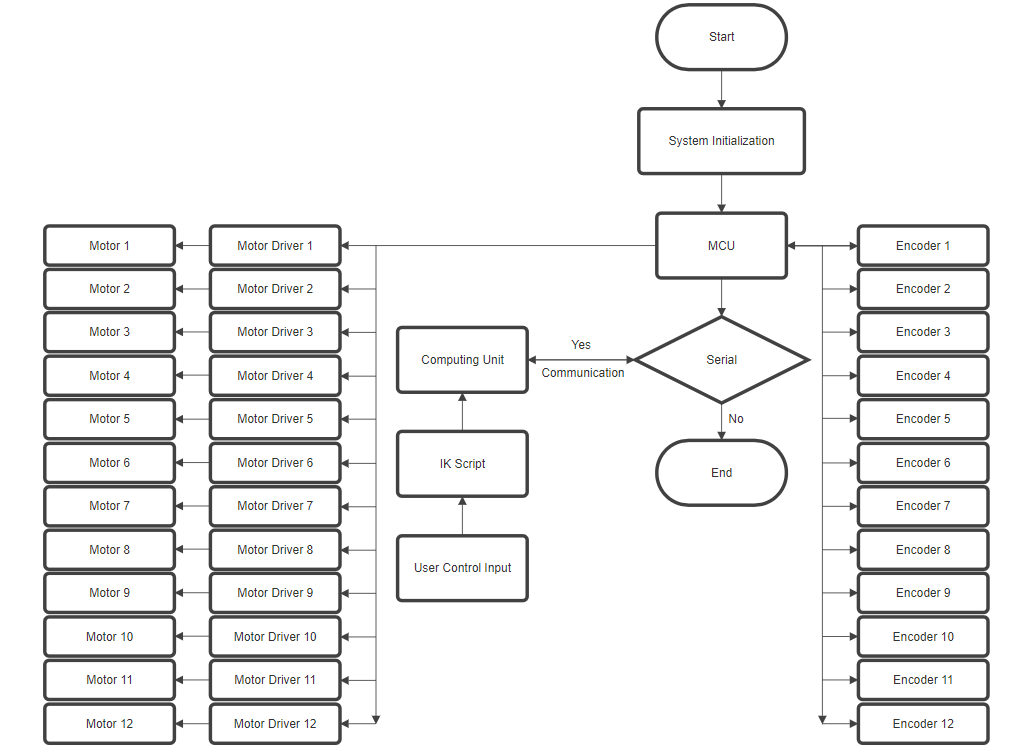}
    \caption{Flow chart of the working principle}
    \label{fig:Flow chart of the working principle}
\end{figure}
\FloatBarrier 

\section{Validation and characterization}
\subsection{PDN Analysis, Optimization and Validation:}
While designing the circuit, we adhered to IPC-A-600H (Acceptability of Printed Boards), IPC-6012D (Qualification and Performance Specification for Rigid Printed Boards), IPC-4101D (Specification for Base Materials for Rigid and Multilayer Printed Boards), and IPC-SM-840E (Qualification and Performance of Permanent Solder Mask) standards for fabricating the PCB hardware. Additionally, we followed the IPC-2221 standard and referenced Yi Wang et al.'s research \cite{ref23} for determining the cross-sectional area of a conductor necessary to handle the required current density. This research is illustrated in Figures 17(a) and 17(b), showing the relationship between the cross-sectional area and the conductor width.
\begin{figure}[h!]
    \centering
    \begin{minipage}{0.49\textwidth}
    \centering
    \includegraphics[width=6cm]{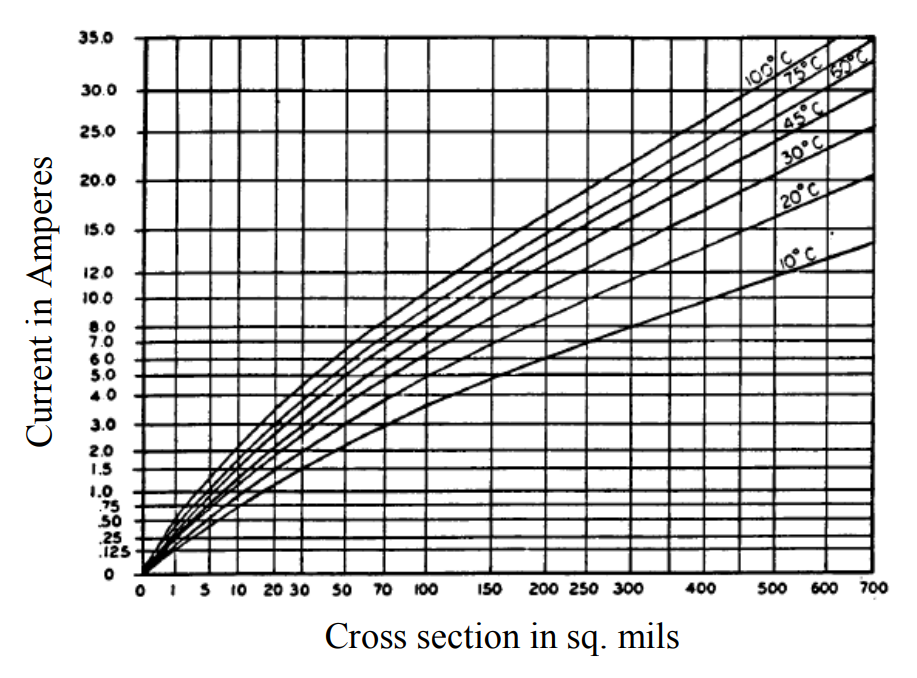}
    \label{fig:(a))}
    \end{minipage}
    \hfill
    \begin{minipage}{0.49\textwidth}
    \centering
    \includegraphics[width=6cm]{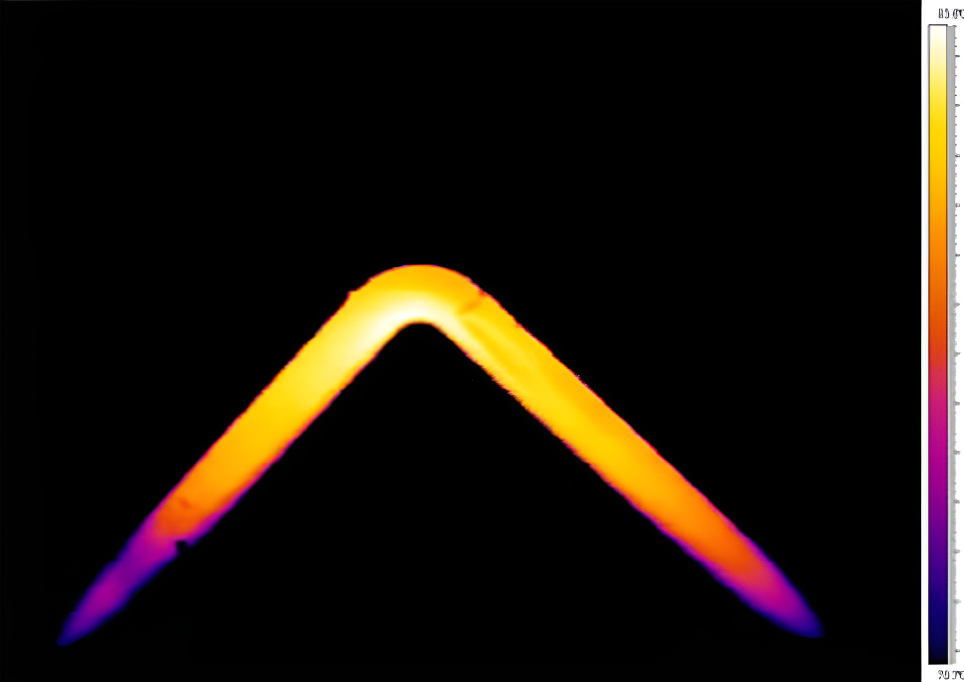}
    \label{fig:(b))}
    \end{minipage}
    \caption{(a) Current density (b) Infrared thermal picture of 90° bended 10mm wide 35µm thick external trace (18A DC current temperature range: 71°C to 83.5°C) \cite{ref26} }
\end{figure}
\FloatBarrier 

The trace width is calculated from the formula:

\begin{equation}
\text{Area} \ [\text{mils}^2] = \frac{\text{Current} \ [\text{Amps}]^k \cdot \text{TempRise} \ [\text{°C}]^b}{c}
\end{equation}

\begin{equation}
\text{Width} \ [\text{mils}] = \frac{\text{Area} \ [\text{mils}^2]}{\text{Thickness} \ [\text{oz}] \times 1.378 \ [\text{mils/oz}]}
\end{equation}

For IPC-2221 internal layers: \( k = 0.024 \), \( b = 0.44 \), \( c = 0.725 \)

For IPC-2221 external layers: \( k = 0.048 \), \( b = 0.44 \), \( c = 0.725 \)

Where \( k \), \( b \), and \( c \) are constants derived from curve fitting to the IPC-2221 curves from Figure 17(a), we have limited our motor control driver to a maximum of 2A. The copper weight is 2 oz/ft², with an ambient temperature of 25°C and a maximum temperature rise of 10°C. This results in a minimum copper trace width of 15 mils (thousandths of an inch) (rounded value) for the significant current-carrying peripherals on the outer layer since the PCB is designed with only 2 layers, eliminating the need to account for inner layer currents.

To compensate, we have routed all the designed tracks with a minimum of 15 mil, and heavy current-carrying traces with 25 mil, which is 60\% more width than necessary to handle the current through the conductors. We have done a Power Density Network (PDN) analysis and detailed information is available in the load performance summary. This allows us to understand and analyze the different sections of the board's power distribution and ensure the design requirements are met.

To verify this, we have conducted a PDN analysis on our designed board, and the visual interpretation of the board voltage drop (see Fig.~18) with current density analysis (see Fig.~19) is done to verify the board’s compatibility with the designed hardware. The PDN analysis result of this board works similarly to a heat map across the board indicating the trace line’s carrying voltage. This analysis is done considering the voltage drop of the accounting peripherals voltage source, converted voltage source, and what are the actual voltage drops happening after the peripherals are under load. The heat map of the voltage (see Fig.~18) illustrates the voltage values across the board. The main trace that is carrying the highest 12V to the motor drivers experienced slight voltage fluctuation, as we can notice in distant connected peripherals. While we maintain signal integrity in trace layout, the reduced voltage happens in this trace due to reflection and trace line impedance. The upper part motors experience slightly lower voltage since the trace line routing has a sharp bend in the layout, which causes signal reflection. As the distance from the voltage source increases, the impedance of the trace line also increases. This higher impedance leads to a voltage drop along the trace line, which is pronounced in peripherals located further from the power source, particularly in DV9, DV7, DV11, DV8, and DV10’s connected nets compared to other connected peripherals. The provided PDN report indicates that despite some voltage drops, the PDN analysis passed, meaning the voltage levels remain within acceptable limits for the design. Table~2 below details the pin voltage and current measurements for major power-distributing loads on the board. The 5V (indicated in blue) and the 9V (indicated in green) remain evenly distributed along the power networks and these networks do not experience any potential voltage drop across the board as these signals connected peripherals consume a very low amount of current and, we have used an LDO (Low-Dropout Regulator). In our layout (see Fig.~19), certain regions exhibit an unacceptably high current density (max 345 A/m²). This increasing amount of current heavily affects the vias, pads, and interconnected tiny trace lines. To minimize this excessive current entry, we have used a ground polygon in the next iteration to evenly distribute the current density and reduce the current density, distributing it over a larger surface area. This resulted in the current-carrying regions staying under minimal current density as we have observed in the simulation that the significant regions (see Fig.~20) stay under a commendable 1.25 to 13 A/m². This reduces the trace’s longevity and prevents burnout. This validation is sufficient to ensure that signal integrity and power plane optimization are evenly addressed to meet the requirements.
\begin{figure}[h!]
    \centering
    \includegraphics[width=7cm]{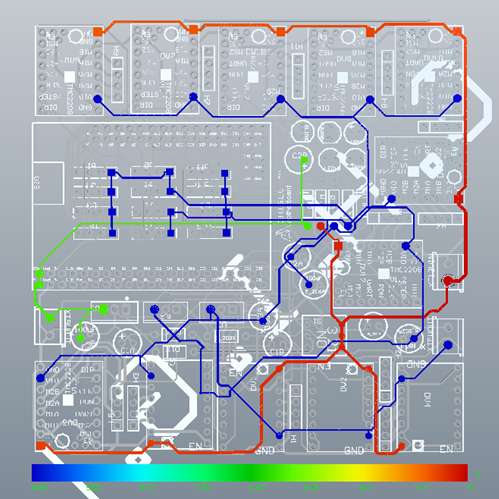}
    \caption{PDN Analyzer: Voltage Analysis}
    \label{fig:PDN Analyzer: Voltage Analysis}
\end{figure}
\FloatBarrier 

\begin{table}[ht]
\caption{Load Performance Summary}
\centering
\begin{adjustbox}{width=\textwidth, center}

\renewcommand{\arraystretch}{3} 
\begin{tabular}{|l|l|l|l|l|l|l|l|l|l|l|l|l|l|}
\hline
\small ID & \small Load Type & \small Ref Des & \small Pass / Fail & \small Load Value & \small Min Voltage (V) & \small Actual Voltage & \small Margin (V) & \small Margin (\%) & \small Min Pwr Pin & \small Voltage (V) & \small Max Gnd Pin & \small Voltage (V) & \small Actual Current (A) \\
\hline
\small Load 1 & \small Current & \small DV1 & \small PASS & \small 1.3A & \small 10.5 & \small 11.032 & \small 532.02m & \small 5.07\% & \small DV1-16 & \small 11.67 & \small DV1-9 & \small 637.48m & \small 1.3 \\
\small Load 2 & \small Current & \small DV2 & \small PASS & \small 1.3A & \small 10.5 & \small 11.04 & \small 1.0397 & \small 10.40\% & \small DV2-16 & \small 11.707 & \small DV2-9 & \small 667.51m & \small 1.3 \\
\small Load 3 & \small Current & \small DV3 & \small PASS & \small 1.3A & \small 10.5 & \small 10.906 & \small 405.66m & \small 3.86\% & \small DV3-16 & \small 11.559 & \small DV3-15 & \small 653.74m & \small 1.3 \\
\small Load 4 & \small Current & \small DV4 & \small PASS & \small 1.3A & \small 10.5 & \small 10.973 & \small 473m & \small 4.50\% & \small DV4-16 & \small 11.676 & \small DV4-9 & \small 702.7m & \small 1.3 \\
\small Load 5 & \small Current & \small DV5 & \small PASS & \small 1.3A & \small 10.5 & \small 10.873 & \small 373.45m & \small 3.56\% & \small DV5-16 & \small 11.528 & \small DV5-1 & \small 654.75m & \small 1.3 \\
\small Load 6 & \small Current & \small DV6 & \small PASS & \small 1.3A & \small 10.5 & \small 11.602 & \small 1.1015 & \small 10.49\% & \small DV6-16 & \small 11.852 & \small DV6-9 & \small 250.16m & \small 1.3 \\
\small Load 7 & \small Current & \small DV7 & \small PASS & \small 1.3A & \small 10.5 & \small 10.818 & \small 318.4m & \small 3.03\% & \small DV7-16 & \small 11.393 & \small DV7-1 & \small 574.2m & \small 1.3 \\
\small Load 8 & \small Current & \small DV8 & \small PASS & \small 1.3A & \small 10.5 & \small 11.062 & \small 561.8m & \small 5.35\% & \small DV8-16 & \small 11.517 & \small DV8-1 & \small 455.4m & \small 1.3 \\
\small Load 9 & \small Current & \small DV9 & \small PASS & \small 1.3A & \small 10.5 & \small 10.758 & \small 258.04m & \small 2.46\% & \small DV9-16 & \small 11.365 & \small DV9-1 & \small 607.27m & \small 1.3 \\
\small Load 10 & \small Current & \small DV10 & \small PASS & \small 1.3A & \small 10.5 & \small 11.169 & \small 669.03m & \small 6.37\% & \small DV10-16 & \small 11.617 & \small DV10-1 & \small 447.87m & \small 1.3 \\
\small Load 11 & \small Current & \small DV11 & \small PASS & \small 1.3A & \small 10.5 & \small 10.949 & \small 448.6m & \small 4.27\% & \small DV11-16 & \small 11.441 & \small DV11-1 & \small 492.5m & \small 1.3 \\
\small Load 12 & \small Current & \small DV12 & \small PASS & \small 1.3A & \small 10 & \small 11.415 & \small 1.4147 & \small 14.15\% & \small DV12-16 & \small 11.734 & \small DV12-1 & \small 319.3m & \small 1.3 \\
\hline
\end{tabular}
\end{adjustbox}
\label{tab:load_testing_results}
\end{table}

\FloatBarrier 

\begin{figure}[h!]
    \centering
    \includegraphics[width=7cm]{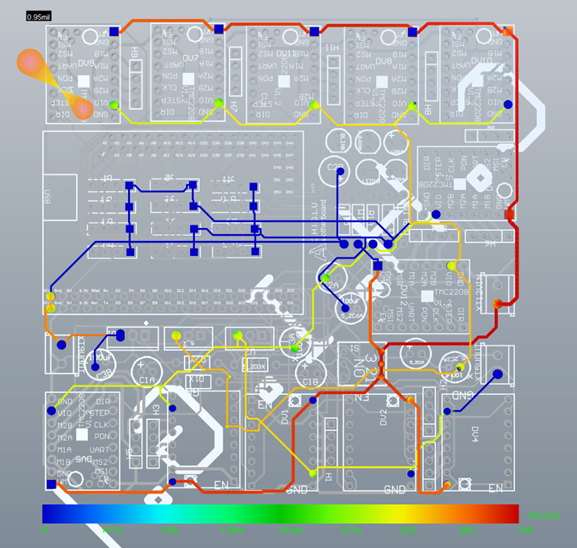}
    \caption{PDN Analyzer: Current Density Analysis}
    \label{fig:PDN Analyzer: Current Density Analysis}
\end{figure}
\FloatBarrier 

\begin{figure}[h!]
    \centering
    \includegraphics[width=7cm]{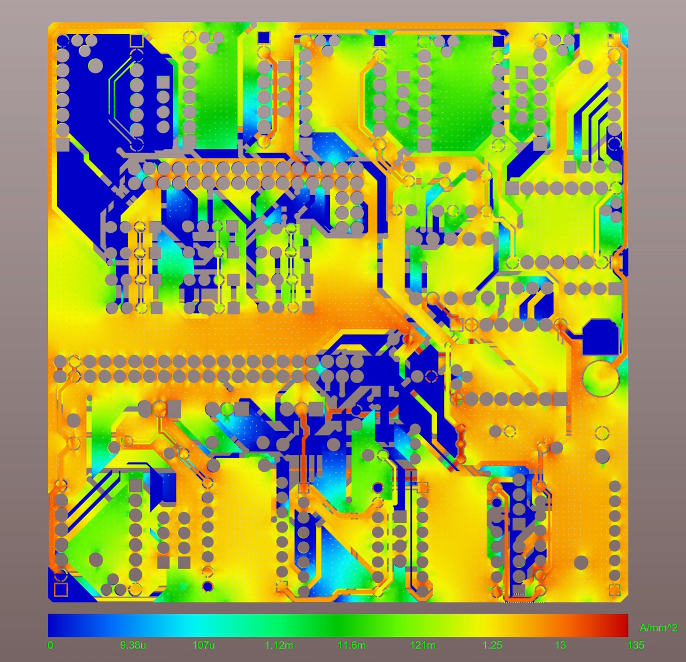}
    \caption{PDN Analyzer: Improved Current Density}
    \label{fig:PDN Analyzer: Improved Current Density}
\end{figure}
\FloatBarrier 

\subsection{Design for Manufacturing}
The primary concentration of this work has been to make the design with accordance to ease with manufacturing process. As per the design requirement, it has been a densely populated layout design. The complexity of the layout could have been reduced if the design could have been shifted to four layers design, where power and signal plane could have been isolated, however this would lead to more manufacturing complexity with increased cost, eventually leading to difficult manufacturing. For this reason, we accomplished this design in two layers, which met our design for manufacturing goal. 

DFM (Design for Manufacturing) concentrates on designing a product from the outset to streamline its manufacturing process, that results in a superior product at reduced costs. To meet this, we have integrated DFM principles early in the project's inception phase. We have used CAM (Computer-Aided Manufacturing) software to inspect and make the design optimized for fabrication. In section 3 of Design files summary, we have provided these CAM files for layers which have been optimized and made design ready for fabrication. 

We have ensured the following DFM Principles:

\begin{itemize}[label=--]
    \item The trace lines are ensured to have a minimum 0.15mm clearance with corresponding traces and polygons to ensure that the manufacturing process can produce a circuit board where traces might not be prone to being shorted (See Fig. 22(e))
    \item The annular rings are checked (See Fig. 22(f)) to ensure that the copper and masking over the rings are above 0.3mm. These minimum requirements ensure that the rings are adequately open to copper and, when subjected to the HASL (Hot Air Solder Leveling) treatment, there are just enough conducting areas where solder can be added.
    \item The drill holes are optimized (See Fig. 22(d)) to meet the minimum number of drill bits required in the drill process. This is achieved by optimizing closely matched drill sizes to one drill size, which reduces the need for multiple drill sizes and ensures hole size compatibility with the components. Both plated and non-plated holes are optimized and made sure that exposed areas are covered with mask (See Fig. 22(c)). Vias are optimized, and we chose to make the vias buried vias, which ensures unnecessary exposing and optimized HASL usage on the PCB.
    \item The board outline is reconstructed to ensure that the rout program maintains a chained path (See Fig. 22(b)). This ensures lesser process time and that the outline is straight rather than curved while cutting down the board.
    \item Silk-screens are optimized to ensure the text’s visibility.
    \item Mask layers are optimized (See Fig. 22(c)) to meet manufacturing requirements. We have increased the mask layer size to 0.05mm to ensure that the copper layer is properly exposed. This also ensures that the hot air solder is exposed to a larger surface area.
\end{itemize}

\begin{figure}[h!]
    \centering
    \includegraphics[width=10cm]{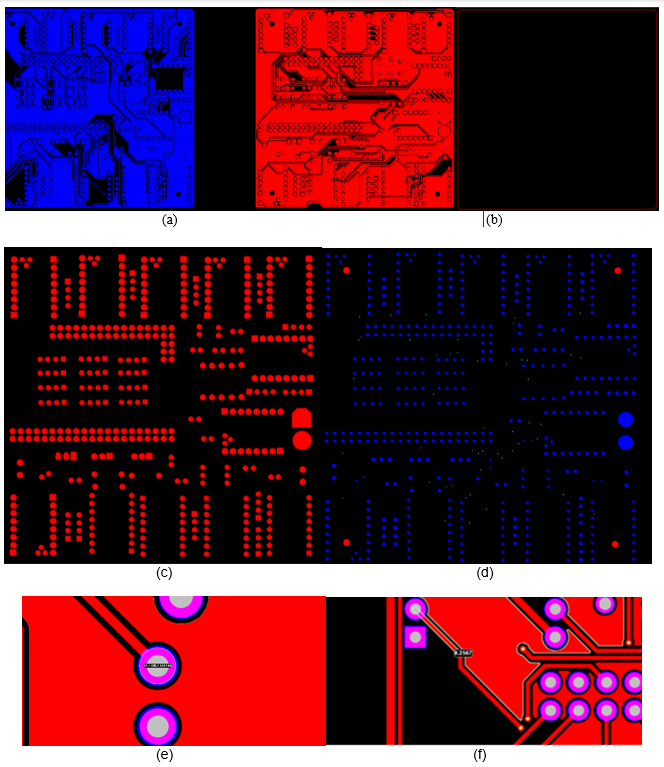}
    \caption{(a) Circuit-layer optimization (b) Outline chained for optimal routing (c) Drill hole optimization (d) Mask layer optimization (e) Trace line clearance adjustment (f) Annual ring size optimization}
    \label{fig:(a) Circuit-layer optimization (b) Outline chained for optimal routing (c) Drill hole optimization (d) Mask layer optimization (e) Trace line clearance adjustment (f) Annual ring size optimization}
\end{figure}
\FloatBarrier 

Along with manufacturing files, this study provides industry-ready AOI (Automatic Optical Imaging) files (see Section 3, Design Files Summary) that could be used to enhance product quality by efficiently identifying defects during the manufacturing process. Additional G-code files are provided with this study for drilling and routing purposes.

\subsection{Hardware Testing and Validation}

\begin{figure}[h!]
    \centering
    \includegraphics[width=\textwidth]{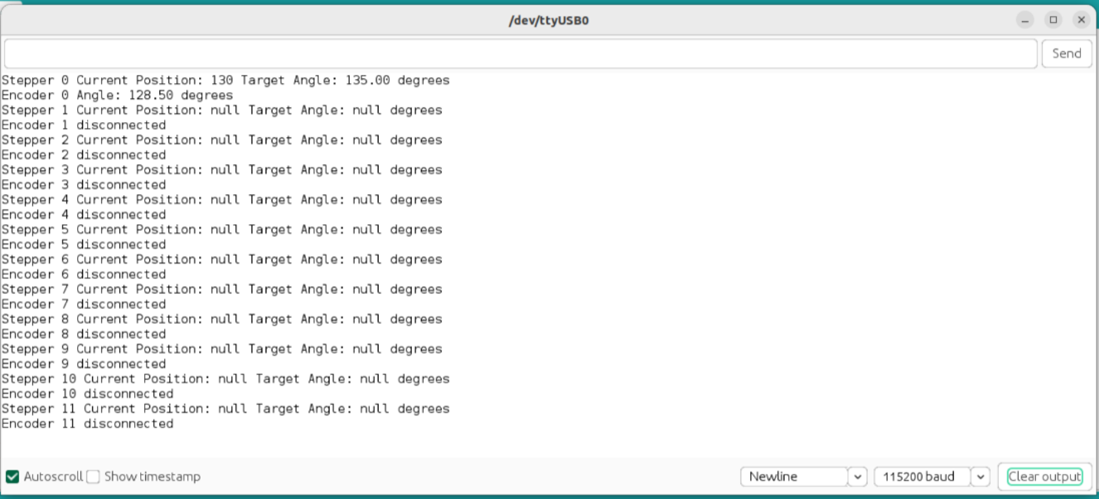}
    \caption{Arduino Terminal Monitor output}
    \label{fig:Arduino Terminal Monitor output}
\end{figure}

To verify with hardware, we set up the full board, stacking all the peripherals and uploaded the Arduino sketch (see Fig. \ref{fig:Arduino Code Snapshots}) to the micro-controller. This setup allows the micro-controller to receive the angle sent from the computing unit through serial communication and interact with the connected peripherals. The provided sketch is designed to work with twelve stepper motors and twelve encoders. For testing and validation purposes, we tuned the code and ran it with one stepper motor and one encoder, receiving the desired response through the serial monitor (see Fig. \ref{fig:Arduino Terminal Monitor output}). To validate that each motor driver functions correctly, we performed a manual voltage test with a multimeter (see Fig. \ref{fig:Bare board Signal Test with Multimeter}) and connected the bare board to a power supply. This process ensures that the hardware has passed the bare board testing.

\subsubsection{Code Functionality}

\begin{figure}[h!]
    \centering
    \includegraphics[width=\textwidth]{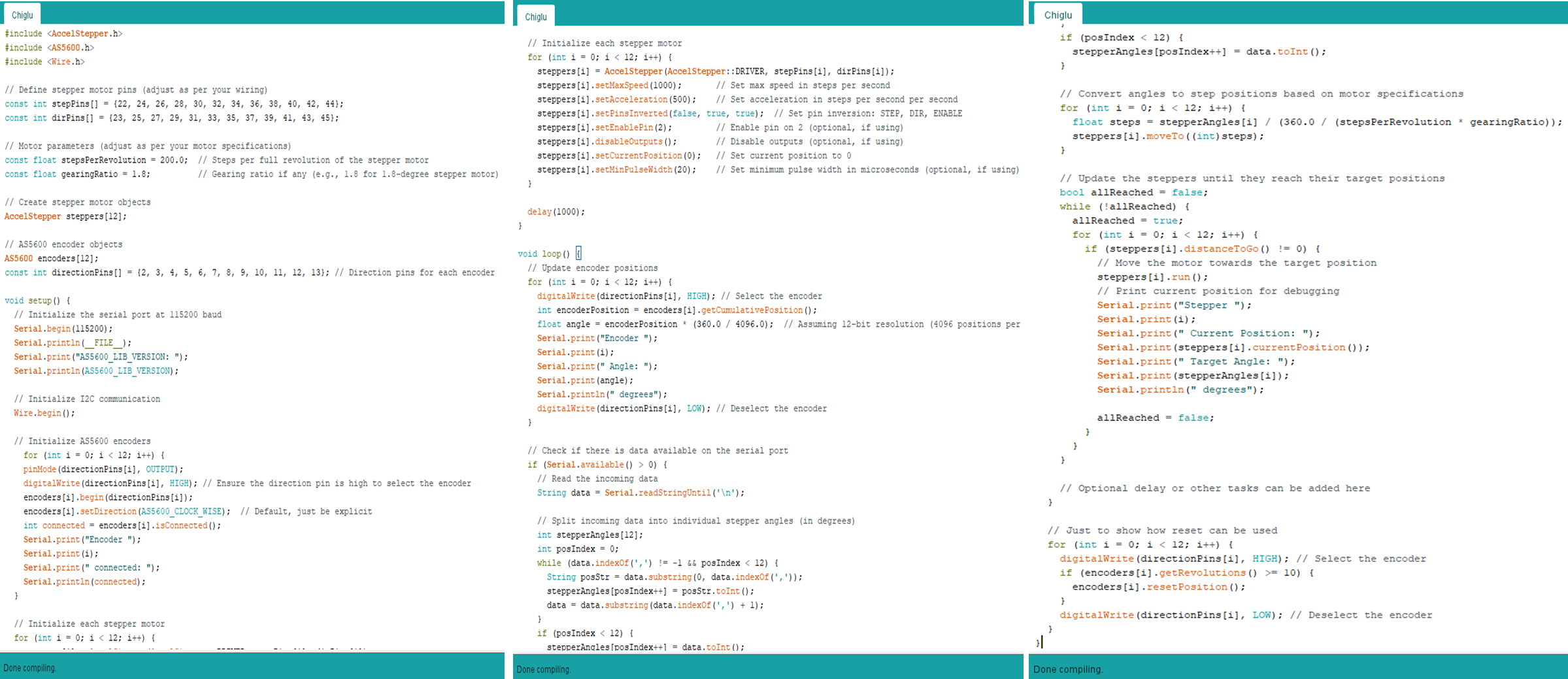}
    \caption{Arduino Code Snapshots}
    \label{fig:Arduino Code Snapshots}
\end{figure}

\begin{figure}[h!]
    \centering
    \includegraphics[width=10cm]{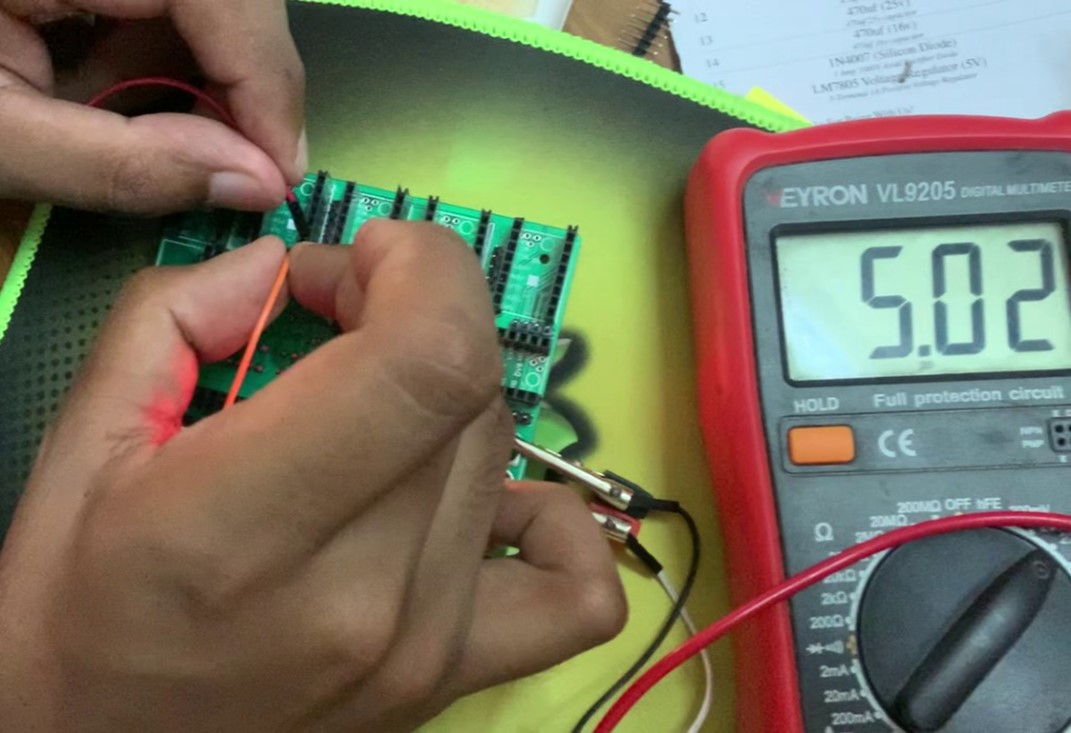}
    \caption{Bare board Signal Test with Multimeter}
    \label{fig:Bare board Signal Test with Multimeter}
\end{figure}

The provided Arduino sketch (see Fig. \ref{fig:Arduino Code Snapshots}) uses the \texttt{AccelStepper} and \texttt{AS5600} libraries to control 12 stepper motors and read positions from twelve AS5600 encoders. Below is a detailed explanation of the code:

\begin{enumerate}
    \item \textbf{Initialization:}
    \begin{itemize}
        \item The \texttt{setup()} function initializes serial communication at a baud rate of 115200 and prints the file name and AS5600 library version.
        \item I2C communication is initialized using the \texttt{Wire.begin()} function.
        \item Each of the 12 AS5600 encoders is initialized. The direction pin for each encoder is set high to select the encoder, and the encoder is set to work in a clockwise direction. The connection status of each encoder is checked and printed to the serial monitor.
        \item Each of the 12 stepper motors is configured with parameters such as maximum speed, acceleration, pin inversion settings, enable pin, current position, and minimum pulse width.
    \end{itemize}

    \item \textbf{Main Loop:}
    \begin{itemize}
        \item The \texttt{loop()} function continuously updates the encoder positions, calculates the corresponding angles, and prints these angles to the serial monitor.
        \item If data is available on the serial port, it reads the incoming data string, splits it into individual stepper angles, and converts these angles to step positions. The stepper motors are then moved to these target positions.
        \item The current position and target angle of each stepper motor are printed to the serial monitor until all motors reach their target positions.
        \item A mechanism is included to reset the encoder position if it completes 10 revolutions.
    \end{itemize}
\end{enumerate}

\subsubsection{Output Example}

Here is an example of the expected output when the sketch is run.

\begin{verbatim}
Stepper 0 Current Position: 130 Target Angle: 135.0 degrees
Encoder 0 Angle: 128.5 degrees
Stepper 1 Current Position: 256 Target Angle: 120.0 degrees
Encoder 1 Angle: 257.3 degrees
... ... ...
... ...
Stepper 11 Current Position: 360 Target Angle: 165.0 degrees
Encoder 11 Angle: 180.0 degrees
\end{verbatim}

\subsubsection{Explanation}

\begin{itemize}
    \item The sketch prints the current position and target angle for each stepper motor in degrees.
    \item The encoder angles are printed to provide feedback on the motor's actual position.
\end{itemize}
\FloatBarrier 

\FloatBarrier 
By running this sketch, we can observe the output on the serial monitor, which includes the current position and target angle of each stepper motor, as well as the angle of each encoder. This helps in verifying the proper functioning of the stepper motor and encoder setup. Positive results from these tests confirm that the system accurately receives and processes angle data and drives the stepper motors accordingly. This validation process, combined with manual voltage testing using a multi meter and power supply, ensures that the hardware setup is working as expected. The system's ability to read encoder positions and control stepper motors based on the received angles demonstrates the expected result from our designed hardware.
\FloatBarrier 
\FloatBarrier 
\section{Conclusion}
This work introduces a low-cost, open-source, nearly fully do-it-yourself, easy replicable electrical hardware system for a stepper motor-based quadruped robot. Recently, the development of quadruped robots and their capabilities have garnered significant attention from developers to experiment with various approaches. The study contributes to the development of introducing an open-source stepper motor-based electrical hardware system both by addressing the very recent study’s research gap and following a hardware system development of a bio-inspired robot with a non-conventional motor system. The paper validates the hardware’s design approach. This study is important because it sheds light on any researcher of a starting point on the design, and architecture of the electrical system by providing detailed power budget planning, schematic and layout, detailed replication instruction, and provides insights on further improvements of the design and extension. The PDN simulation and hardware testing passed the requirements and can be used as an electrical control system for a complete quadruped robot. Preliminary results acquired from the PDN analyzer have shown that the voltage and current density parameters are within the desired range and passed the requirement and we have tested manually with probes in bare board PCB testing after fabrication and checked each voltage point. However, due to limited access to multiple motor drivers and stepper motors the hardware testing has been limited to one stepper motor and one encoder, but we have tested it in every twelve driver and encoder out headers. Although with this limitation in the hardware testing, the PDN analyzer report provides validation on the performance of the designed hardware under full load. This paper also provides design optimization in PDN analysis providing insights for researchers and developers on how the design optimization has been made, thereby giving insight to future developers and researchers on how they can further achieve development. We have deployed the code to the MCU and conducted the testing of the function. We have been successful in moving the motor with generated angles, which proves successful communication through the designed PCB boards and computing unit. Also, we received feedback from the encoders which demonstrates that our developed hardware is working as expected. Our developed hardware meets the design requirement and future works with this hardware can open endless improvement possibilities with the fully built physical robot. The research community can start from where this literature ended.

\section*{Inspiration for the Project}

I, Abid Shahriar, as the first author, with the consent of my co-author, would like to dedicate this work to the memory of Chiglu, the beloved cat that my sister, Fariha Zabin Puspita, and I shared. The joy and companionship the pet had brought to our lives have been a significant influence during the COVID-19 pandemic, inspiring this exploration into bio-inspired robotics. This project is a reflection of the love and inspiration Chiglu brought into our lives and stands as a testament to our drive to make meaningful contributions to bio-inspired robots and beyond.

\FloatBarrier 


\end{document}